# Combining Kernelized Autoencoding and Centroid Prediction for Dynamic Multi-objective Optimization

Zhanglu Hou, Juan Zou∗ , Gan Ruan, Yuan Liu, Yizhang Xia

**Abstract**    Evolutionary algorithms face significant challenges when dealing with dynamic multi-objective optimization because Pareto optimal solutions and/or Pareto optimal fronts change. This paper proposes a unified paradigm, which combines the kernelized autoncoding evolutionary search and the centroid-based prediction (denoted by KAEP), for solving dynamic multi-objective optimization problems (DMOPs). Specifically, whenever a change is detected, KAEP reacts effectively to it by generating two subpopulations. The first subpopulation is generated by a simple centroid-based prediction strategy. For the second initial subpopulation, the kernel autoencoder is derived to predict the moving of the Pareto-optimal solutions based on the historical elite solutions. In this way, an initial population is predicted by the proposed combination strategies with good convergence and diversity, which can be effective for solving DMOPs. The performance of our proposed method is compared with five state-of-the-art algorithms on a number of complex benchmark problems. Empirical results fully demonstrate the superiority of our proposed method on most test instances.

## 1  Introduction

There are many real-world dynamic multi-objective optimization problems (DMOPs), such as software project scheduling [1] [2], big data optimization [3] [4], resource management [5] [6], whose objective functions, constraints or variables may change over time. Due to the property of time dependent variability, it is challenging to quickly track the moving Pareto optimal set (POS) and maintain the

∗  Corresponding author

Z. Hou, J. Zou, Y. Liu, and Y. Xia are with Hunan Engineering Research Center of Intelligent System Optimization and Security, Key Laboratory of Intelligent Computing and Information Processing, Ministry of Education of China, and Key Laboratory of Hunan Province for Internet of Things and Information Security, Xiangtan University, Xiangtan, 411105, Hunan Province, China E-mail: ahou.amstrong@gmail.com, Zoujuan@xtu.edu.cn, LiuYuan@xtu.edu.cn, Yizhang@xtu.edu.cn, (Corresponding author: Juan Zou)

G. Ruan is with CERCIA, School of Computer Science, University of Birmingham, Edgbaston Birmingham B15 2TT, UK. E-mail: GXR847@cs.bham.ac.uk.

S. Yang is with the Centre for Computational Intelligence, School of Computer Science and Informatics, De Montfort University, Leicester, LE19BH, U.K. E-mail: syang@dmu.ac.uk

diversity of population. In recent years, more and more researchers are devoting themselves to the research of dynamic multi-objective evolutionary algorithms (DMOEAs). Especially, a significant amount of research has been dedicated to response strategies. If the environment is deemed changed, the response mechanism can be carried out via tracing the moving POS in the changing environment. Specifically, the change response techniques can be classified into four main categories. 1) diversity-based approaches. 2) memory-based approaches. 3) multi-population approaches. 4) prediction-based approaches. We will briefly introduce these four methods in the following section and the interested reader is referred to [7] for a recent comprehensive survey.

Diversity-based approaches usually introduce extra diversity or maintain high diversity for a new environment, trying to rescue the diversity loss caused by environmental changes [8]. There are two commonly used approaches in the literature for introducing diversity into a population when a change is detected. The first approach involves immediately increasing the population by introducing randomly solutions, while the second approach involves hyper-mutating some historical solutions in the population. Building upon the nondominated sorting genetic algorithm II (NSGA-II) [9], Deb et al. [10] integrated these two approaches into the NSGA-II framework to achieve diversity increase. However, diversity enhancing mechanisms may not be effective in tackling more complex DMOPs. This is because promoting excessive diversity can have a negative impact on search efficiency, leading to slower convergence speeds [11].

Utilizing memory pools to store high-quality solutions from past environments is referred to as memory-based approach [12] [13]. The solutions stored in the memory pool are then reused as initial solutions for a new environment, which can help to accelerate convergence. However, the effectiveness of memory-based approaches is heavily influenced by the similarity between different environments. When addressing DMOPs that exhibit non-periodic characteristics, the effectiveness of memory-based approaches is greatly reduced [14]. In addition, a certain amount of computation is required to store historical information in memory [15].

Multi-population approaches have advantages to mitigate the loss of diversity during optimization [16], which involves utilizing multiple subpopulations distributed throughout the search process. These subpopulations can interact with each other in a competitive or cooperative manner to help maintaining diversity [17] [18]. Moreover, this method is particularly efficient when certain search areas undergo modifications in a new environment while others remain unchanged [19].

Prediction-based approaches have gained significant popularity for solving DMOPs, partly because the problems of two adjacent time periods generally have similarities, rather than being totally different from each other [20] [21]. This kind of prediction approaches mainly building a linear or nonlinear model can leverage previous search information to predict DPOS in the new environment for DMOPs with predictable changes. In the past few years, there has been a growing interest in developing advanced prediction methods to track dynamic changes in order to efficiently and effectively solve DMOPs.

Recently, more and more machine learning techniques are being introduced to help solving DMOPs [22] [23]. In particular, autoencoding evolutionary search is proposed to guide the evolution of population by utilizing knowledge gained from past search experiences [24]. This kind of search paradigm relies on a learning component that utilizes a single layer denoising autoencoder (DA). This DA is a modified version of the traditional autoencoder and has a closed-form solution, which makes it computationally efficient for use with the evolutionary solver [25]. In [25], the prediction method being proposed consists of two main components: a linear autoencoding-based prediction (AE) and preservation of high quality solutions. However, despite the success applied by autoencoding evolutionary search for solving DMOPs, the linear autoencoding model cannot capture the nonlinear relationship between the historical solutions used in the mapping construction [26]. Meanwhile, numerous linear or nonlinear prediction models have been designed to forecast environmental change patterns, but we are still uncertain whether DPOS follows a linear or non-linear variation when the environmental changes occur. To track the dynamic DPOS more effectively, this paper presents a new prediction strategy that combines the kernelized autoencoding (KAE) and centroid-based prediction strategy, called KAEP, where the more complex changes of historical solutions can be captured by mapping the historical POS into a reproducing kernel Hilbert space based the kernel method. To summarize, the main contributions of this paper are presented as follows:

1) To cope with complex environmental changes in DOMPs, we have proposed a unified paradigm that trying to combining a linear prediction model and a non-linear prediction model. Specifically, we propose a novel approach of responding to changes in DMOPs via the KAE evolutionary search and a simple centroid-based prediction. With the learned nonlinear mapping, the knowledge from past searching POS experience can provide more diverse and accurate POS prediction for solving DMOPs with complex characteristics.

2) To verify the efficacy of the proposed method, comprehensive empirical studies have been conducted on the commonly used DMOP benchmarks that possess various characteristics. The obtained results confirmed the effectiveness of the proposed KAE evolutionary search and centroid-based prediction for solving DMOPs. Additionally, an ablation study is conducted to verify that KAE indeed outperforms AE when using the same static optimization algorithm.

The rest of this paper is organized as follows. The background and related research, including the related definitions of DMOPs, change detection and related works, are introduced in Section 2, respectively. Then, Section 3 details the proposed KAEP method and experimental results and analysis are given in Section 5. Finally, the conclusion of this paper is given in Section 6.

## 2 Background and Related Research

### 2.1 The related definitions of dynamic multi-objective optimization

In this section, we will give some definitions of dynamic multi-objective optimization. Without loss of generality, for a minimization problem, the DMOP is mathematically defined as follows:

$$\begin{cases} minF(x,t) = (f_1(x,t), f_2(x,t), ..., f_m(x,t))^T \\ s.t. \ g_i(x,t) \geq 0, \ i = 1,...,n_g \\ \qquad h_j(x,t) = 0, \ j = 1,...,n_h \end{cases} \qquad (1)$$

Where $x = (x_1, x_2, ..., x_n) \in \mathcal{R}^n$ is the decision vector which consists of n decision variables, and $F(x,t): \mathbb{R}^n \times t \to \mathbb{R}^m$ is the objective vector which consists of m time-varying objective functions. $g_i$ is the ith inequality constraint and $h_j$ is the jth equality constraint. $n_h$ and $n_g$ are the number of equality constraints and inequality constraints, respectively.

Definition 1: (Pareto Dominance) At time t, a decision vector $x_1$ is said to Pareto dominate another decision vector $x_2$, denoted by $x_1 \prec_t x_2$, if and only if

$$\begin{cases} \forall i \in \{1, ..., m\} & f_i(x_1, t) \leq f_i(x_2, t) \\ \exists i \in \{1, ..., m\} & f_i(x_1, t) > f_i(x_2, t) \end{cases} \qquad (2)$$

The definitions of the dynamic Pareto-optimal set (DPOS) and dynamic Pareto-optimal front (DPOF), derived from the concept of Pareto dominance, are given as follows.

Definition 2: (DPOS) At time t, $x^*$ is identified as a Pareto optimal solution if there do not exist x such that $x \prec_t x^*$, then the set including all Pareto optimal solution $x^*$ is defined by

$$DPOS = \{x^* | \nexists x \in \Omega, x \prec_t x^*\} \qquad (3)$$

Definition 3: (DPOF) At time t, the corresponding objective vectors of DPOS form the dynmaic POF (DPOF), denoted by

$$DPOF = \{F(x^*, t) | x^* \in DPOS\} \qquad (4)$$

### 2.2 Change Detection

In this section, we will present some change detection techniques of DMOEAs, which the process of dynamic detection allows an MOEA to adapt to changing environments and maintain optimal performance. Existing change detection methods can be broadly classified into tree categories: random re-evaluation detection, population-based detection and sensor-based detection [7] [27]. Specifically, the random re-evaluation method, the most widely used method for detecting changes in DMOEAs, is to re-evaluate certain random members of the population, as it has gained significant efficiency in the most benchmark problems without uncertainties. Population-based detection methods utilize fitness evaluations of the entire population, while sensor-based detection approaches involve measuring fitness landscapes at specific predefined points. Our proposed method, building on the successes of many previous DMOEAs [28] [29] [30], utilizes a simple and effective reevaluation based detection approach. Specifically, we randomly select 10% of individuals in the population as detectors and archive their objective values. At the beginning of each generation, the detectors are re-evaluated, and a discrepancy in objective values suggests a change in the DMOP.

## 2.3 Autoencoding Evolutionary Search

In this section, we simply introduce a standard autoencoder and its derived applications, especially in multi-objective optimization.

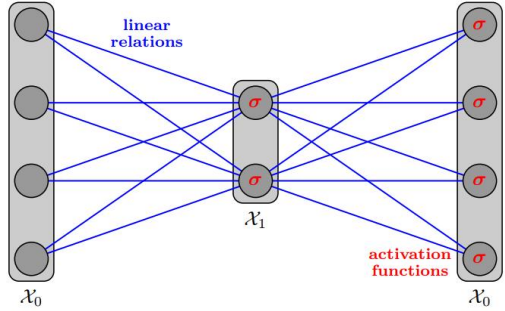

Fig.1. The schematic diagram of a standard autocoder

A standard autoencoder is a type of neural network that is used for unsupervised learning, which means that it can learn patterns in data without being explicitly told what to look for [31]. The autoencoder consists of an encoder that maps the input data into a lower-dimensional representation and a decoder that reconstructs the original data from this lowerdimensional representation [32]. A standard autocoder [33] is shown in 1, suppose that $S = (x_1, x_2, \ldots, x_n) \in \chi_0$ , $\chi_0 = R^d$ and $\chi_1 = R^p$,

$$\begin{cases} f: X_0 \to X_1, f(x) = \sigma(W_1 x + b_1) \\ g: X_1 \to X_0, g(y) = \sigma(W_2 y + b_2) \\ \min_{W_1, W_2, b_1, b_2} \frac{1}{n} \sum_{i=1}^{n} \parallel x_i - g \circ f(x_i) \parallel_{X_0}^2 \end{cases} \qquad (5)$$

where $W_1 \in \mathbb{R}^{P \times d}, b_1 \in \mathbb{R}^p, W_1 \in \mathbb{R}^{P \times d}, b_1 \in \mathbb{R}^d$ . The standard autoencoder is generally to find the linear relations and learn a compressed representation of the input data that captures the most important features [34]. As presented in (5), given the input vector $x \in \chi_0$, a hidden representation $\chi_1$ is captured by a linear mapping $f(x) = \sigma(W_1 x + b_1)$. Beside their conventional use in machine learning for extracting high-quality features [35], the autoencoder has been recently utilized as a way to establish a link between two distinct optimization domains [24] [25]. Instead of simply using the hidden representation $\chi_1$ as a replacement for the original data, they suggest using it as a connection between the corrupted input $\chi_0$ and the repaired "clean" input $\chi_1$. More importantly, the autoencoding evolutionary search allow us to learn historical search experience across different problems and improve the efficiency of evolutionary search in the dynamic context of continuous optimization [25].

Specifically, considering that $S_s = (s_1, s_2, \ldots, s_n) \in \mathbb{R}^{d \times N}$ and $T_t = (t_1, t_2, \ldots, t_n) \in \mathbb{R}^{d \times N}$ are the optimal solution sets from solving two different optimization problems, respectively, where d is the dimensionality of a solution and N is the size of solution sets. The connection $M \in \mathbb{R}^{d \times d}$ of S and T is mathematically constructed by a linear mapping shown as follows.

$$L(M) = \frac{1}{2N} \sum_{i=1}^{N} \parallel t_i - M s_i \parallel^2 \qquad (6)$$

Accordingly, another way to express the information from (6) is to use the closed-form solution [25] [36], which is given as follows.

$$M = (T_t S_s^\top)(S_s S_s^\top)^{-1} \qquad (7)$$

where $\top$ is the transpose operation of a matrix, and suppose that there exists some kind of underlying connection between different problems, one way to transfer knowledge from one problem domain $S_s$ to another $T_t$ is by multiplying a matrix M with the optimized solutions from S.

In addition to seeking the linear relationships among different optimization problems, Zhou et al. [26] proposed a self adaptive Kernelized autoencoding model to capture the linear and nonlinear relationships of heterogeneous problems, and the kernelized autoencoding model is given by:

$$L(M) = \frac{1}{2N} tr\left[(-M\Phi(S_s))^\top (T_t - M\Phi(S_s))\right] \quad (8)$$

where $tr(\cdot)$ denotes the trace of a matrix and $\Phi$ is a nonlinear mapping function [22]. Through the (8), the solution set $S_s$ can be mapped to Kernel Hilbert Space by $\Phi$; meanwhile, according to [26], $M = M_k\Phi(X)^\top$, the kernel matrix can be denoted as $K(S_s, S_s) = \Phi(S_s)^\top \Phi(S_s)$. Finally, a closed-form solution can be rewritten for (8) by deducing it in a another way, which is shown as follows.

$$M_k = T_t K(S_s, S_s)^\top (K(S_s, S_s)K(S_s, S_s)^\top)^{-1} \quad (9)$$

where the (i, j)th element of $K(S_s, S_s)$ is the kernel function value of $\kappa(s_i, s_j)$ and the commonly used polynomial kernel $\kappa(x, y) = (x^\top y + 0.1)^d$ is adopted [37] for nonlinear mapping. Compared to linear autoencoding, the calculation of the kernel matrix is an additional cost. However, this cost is negligible in the context of an evolutionary-search process, and does not significantly increase computational burden.

## 2.4 Related Work

In order to efficiently adapt to dynamic environments and track time-varying DPOS, prediction-based strategies have been proposed. These strategies involve initializing the population and adjusting to environmental changes primarily through a linear or nononlear predictive model [7]. For example, Zhou et al. [38] proposed a population prediction strategy (PPS) that combines the predicted POS centroids with the linear changes of manifold to predict the entire population of DMOPs. However, it faces a limitation in its ability to make accurate predictions during the early stages of evolutionary search because of the lack of sufficient historical information. In [39], the authors developed a grey predictive model to predict population in a new environment. The model uses cluster centroids from previous environments to generate a portion of the initial population in the new environment. Muruganantham et al. [40] proposed the use of the MOEA/D [41] that employs a linear Kalman filter to forecast the POS in a change. The Kalman filter prediction is combined with random reinitialization, using a scoring system, to generate the new population for the change. Rong et al. [42] presented several predictive models for population prediction and introduced a model selection technique. The approach involved identifying the type of change in population size and then choosing the most appropriate predictive model for that particular type of change. Furthermore, Cao et al. [43] proposed a model for predicting the movement of POS by utilizing historical centroid locations. The liner model was designed to calculate the difference between successive centroid locations in order to make predictions about the moving of the POS. Jiang et al. [44] introduced KT-DMOEA, a knee point-based imbalanced transfer learning (nonlinear) model designed to improve the performance in handling DMOPs by transferring knee points. Nevertheless, the time-varying distributions of knee points across different environments pose challenges in accurately predicting knee points in a new environment. In order to mitigate the erroneous prediction caused by track the special points, Yu et al. [45] proposed a correlation-guided layered prediction approach where the integration of multiple linear prediction models takes into account the correlation of individuals' moving directions. In reference [25], for prediction by denoising autoencoding, a single-layer linear autoencoder has been derived to tracking the moving direction of DPOS from the historical nondominated solutions. Similarly, a single autoenndoer model with a nonlinear kernel function [46] has been designed to responding to the environmental changes, where the proposed predictive model aims to learning the knowledge from the search process of historical time steps.

# 3 The Proposed Method

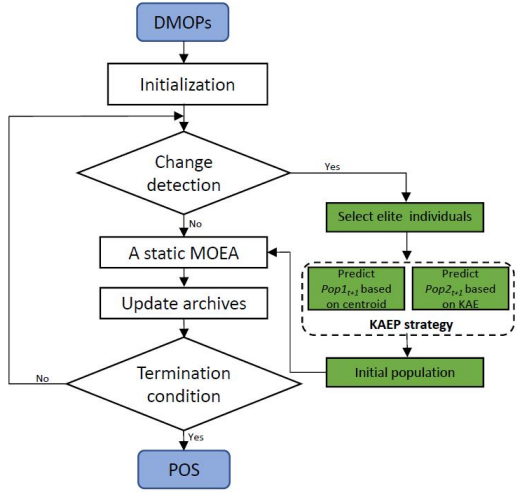

Fig.2. The flowchart of the proposed KAEP-SMOEA for solving DMOPs.

---

**Algorithm 1** The main framework of the KAEP-SMOEA.

**Input:** $SMOEA$ (a static MOEA), $N$ (population size).
**Output:** $POF$ (the approximate POF in different environments).

1: $Pop \leftarrow$ Initialization$(N); t \leftarrow 0;$
2: **while** stopping condition is not met **do**
3:     **if** change detected **then**
4:         $t \leftarrow t + 1$
5:         **if** $t < 2$ **then**
6:             $Pop \leftarrow POS_{t-1}$
7:         **else**
8:             $Pop \leftarrow$ **KAEP**$(Pop_{t-2}, Pop_{t-1})$
9:         **end**
10:     **else**
11:         $POF \leftarrow SMOEA(Pop)$
12:     **end**
13: **end**

---

Considering the limitations of existing DMOEAs leveraging linear autoencoder for prediction when dealing with nonlinear mappings, we propose to utilize kernelized autoencoder (KAE) to do the prediction. Furthermore, to remedy the inaccuracy induced by KAE prediction, we introduce centroid prediction to guide the movement of DPOSs. Therefore, in this section, we propose the strategy of combining KAE and centroid-based prediction (denoted as KAEP) to tackle DMOPs. Specifically, the overall framework of the dynamic algorithm embedding our poposed KAEP is presented in Section III-A. Section III-B describes the specific process of our proposed KAEP strategy.

## 3.1 Overall Framework of Our Proposal

Note that our proposed KAEP is a response strategy to environmental changes and it can be embedded any static multi-objective evolutionary algorithm (SMOEA) to form a DMOEA. We denote the DMOEA embedding KAEP with any SMOEA as KAEP-SMOEA. Figure 2 presents the flow chart of KAEP-SMOEA. Specifically, KAEP-SMOEA starts with an initialization process to randomly initialize a new population. Then, conduct the change detection process to detect whether there are environmental changes. If no, any SMOEA can be adopted to optimize the population; else, our proposed KAEP strategy is evoked to initialize a new population for the new environment.

The framework of the KAEP-SMOEA is presented in Algorithm 1. Given the population size N, KAEP initialize a population with N solutions and set the initial time step as t. Then, lines 2 to 12 are conducted until the stopping condition is met. If an environmental change is detected, as shown in line 3, the time step is increased by 1 in line 4. When the time step t is less than 2, our proposed KAEP will not be adopted since KAEP relies on the DPOSs of previous two environments; instead, we will initialize the population by just coping the previous DPOS (DPOS$_{t-1}$) to $Pop$, as shown in line 6. When the time step is larger than 2, we initialize a population using our proposed KAEP by regarding the population of previous two environments ($Pop_{t-2}$ and $Pop_{t-1}$) as the input, which will be detailed in Section. III-B. If there is no change detected, any SMOEA can be adopted to optimize the population $Pop$ to get the DPOF, as shown in line 11.

## 3.2 Proposed Kernelized Autoencoding and Prediction (KAEP) Strategy

---

**Algorithm 2** The KAE and centroid-based Prediction (KAEP).

**Input:** $C_t$ and $C_{t-1}$: historical centroids obtained in time window t and t-1;

$Pop_t$ and $Pop_{t-1}$: the population solutions obtained in time window t and t-1;

$N$: the size of population.

**Output:** the $initPop$ of new time window t+1.

1: **Begin**
2: $Dt \leftarrow$ calculate the direction via (11)
3: $POS1_t \leftarrow$ environmental selection($Pop_t$, $N/2$)
4: $POS1_{t-1} \leftarrow$ environmental selection($Pop_{t-1}$, $N/2$)
5: $initPop1_{t+1} \leftarrow$ initlize half of population via (12)
6: $M \leftarrow$ obtain $M_k$ with $POS1_{t-1}$ and $POS1_t$ via (9)
7: $initPop2_{t+1} \leftarrow$ generate N/2 solutions via (13)
8: $initPop \leftarrow initPop1_{t+1} \cup initPop2_{t+1}$
9: **End**

---

Since the dynamic environment at two consecutive time steps is different but related, historical search experiences can provide valuable knowledge for future optimization processes [47]. The combination of machine learning and DMOEAs is a promising approach as it enables the reuse of past information about DPOSs or DPOFs from different previous time instances to search the next DPOS in the new environment. To accelerate the search process while improving the quality of the solutions in different environments, we propose a novel KAE-assisted prediction strategy that the KAE is used to predict moving directions of DPOS by leveraging the past search experiences.

Unlike recent many methods [22] [24] that solely rely on machine learning, especially transfer learning, our proposed approach combines simple but widely-used centroid-based prediction with machine learning techniques to hopefully handle complex DMOPs better. The proposed KAEP strategy is described in detail in Algorithm 2. Specifically, supposing that $C_t$ is the centroid of $POS_c$ at time t, and $C_t$ can be expressed as follows.

$$C_t = \frac{1}{|POS_c|} \sum_{x_t \in POS_c} x_t \qquad (10)$$

where $|POS_c|$ is the number of the DPOS of the first ranked NDsort [9] at time window $t$, and $x_t = (x_t^1, x_t^2, \ldots, x_t^n)$ is an individual at time $t$. Thus, the moving direction of centroids referred to as Dt at time $t$ can be calculated in line 2 of Alogrithm 2, where its definition is shown as follows.

$$D_t = C_t - C_{t-1} \qquad (11)$$

To predict the DPOS of a new environment using higher quality optimal solutions and calculate the learned matrix M, we first conduct the environmental selection by crowding distance sorting in [9] for the $Pop_t$ and $Pop_{t-1}$ obtained in the previous environments in line 3 to 4, respectively. Then half of initial population $initPop1_{t+1}$ at time t+1 would be generated in line 5 according to the following formula.

$$x_{t+1} = x_t + D_t \qquad (12)$$

Subsequently, we calculate the $M_k$ introduced in section II-C via (9) in line 6. To transfer a set $POS1_t$ of elite solutions from one problem to the other, we apply the kernel function to get $\Phi(POS1_t)$. Next, we multiply $\Phi(POS1_t)$ by M to obtain the predicted solutions $initPop2_{t+1}$ in line 7, which its specific calculation method is presented as follows.

$$initPop2_{t+1} = M \Phi (POS1_t) = M_k K (POS1_{t-1}, POS1_t) \qquad (13)$$

Finally, two subpopulations in $initPop1_{t+1}$ and $initPop2_{t+1}$ are output in line 8 as the initial population to response a new environment.

## 3.3 Complexity Analysis of KAEP

In this subsection, we analyse the computational complexity of KAEP when applied to a single environmental change. In KAEP, the computational costs are mainly related to the process of environmental selection (line 3 to line 4 of Algorithm 2) and calculating the learned matrix M (line 6 of Algorithm 2 ). Specifically, the environmental selection procedure spends $O(mN^2)$ [9] computation, where m is the number of objectives and N is the size of population, respectively. In order to obtain the value of M, the calculation process costs $O((D+1)^2N)$ computation, where D is the dimensionality of solutions. Thus, the total computational complexity of KAEP is $O(mN^2 + (D+1)^2N)$.

# 4  Experimental Setup

In the following experiments, we validate the proposed KAEP by incorporating it with a widely used MOEA, the NSGA-II [9], named KAEP-NSGA-II. Moreover, we compare KAEP-NSGA-II with five state-of-the-art techniques [10] [22] [25] [46] and the details are introduced in section IV-B. In order to ensure fairness as far as possible, they are also incorporated into NSGA-II, namely, DNSGA-IIA, DNSGA-II-B, PPS-NSGA-II, AE-NSGA-II, KL-NSGA-II, respectively.

In the remainder of this section, a brief introduction of benchmark problems and the compared algorithms is firstly presented. Then we give the parameter settings of each algorithm and also introduction of the adopted performance indicators.

## 4.1  Benchmark Problems

In this section, a total of 14 DF [48] problems with various complicated characteristics are conducted to analyse the performance of the proposed KAEP-SMOEA and compared DMOEAs. The dimension of decision variable is set to 10, and a time variable t in these benchmark problems, which usually control the environmental changes, is defined as follows.

$$t = \frac{1}{n_t}\left\lfloor\frac{\tau}{\tau_t}\right\rfloor \qquad (14)$$

where $n_t$, $\tau_t$ and $\tau$ are the severity of environmental changes, the frequency of environmental changes, and the generation counter, respectively. In this article, according to [11] [49], the severity of change ($n_t$) is fixed to 10, and the frequency of change ($\tau_t$) is set to 5, 10, and 20, respectively.

## 4.2  Compared State-of-the-Arts

To empirically investigate the performance of the proposed prediction strategy, five state-of-the-arts are embedded into NSGA-II and the compared algorithms are briefly introduced as follows.

1) DNSGA-II: DNSGA-II [10] is a dynamic version of NSGA-II by making some changes to the original NSGA-II. Two versions are available: 1) DNSGAII-A and 2) DNSGA-II-B. The former is formed by replacing 10% portion of the population with random solutions and the latter is similar to the former except that the replacement uses mutated solutions of existing individuals

2) PPS: PPS [22] is a representative of prediction-based methods that model the movement track of the DPOF or DPOS in dynamic environments and then use this model to predict the new location of DPOS. In PPS, the DPOS information is divided into two parts: 1) the population centroid and 2) manifold. Based on the archived population centroids over a number of continuous time steps, PPS employs a univariate autoregression model to predict the next centriod in the new environment. Likewise, previous manifolds are used to predict the next manifold. When a change occurs, the initial population for the new environment is created from the predicted centroid and manifold.

3) AE: The AE prediction strategy [25] presents a novel approach to address DMOPs by adopting autoencoding evolutionary search. It consists of two main components: prediction via denoising autoencoding and high-quality solution preservation, which can track the movement direction of the DPOS during the online evolutionary search. In particular, to achieve prediction via denoising autoencoding, the authors derive a closed-form solution for a

CAAI TRANSACTIONS ON INTELLIGENCE TECHNOLOGY

**TABLE I**

The statistics of MIGD results ( mean and standard deviation ) achieved by six compared algorithms on bi-objective and tri-objective DF, where '+', '−', and '≈' indicate each compared algorithm is significantly better than, worse than, and tied by KAEP-NSGA-II, respectively.

| Problem | $(\tau_t, n_t)$ | DNSGA-II-B | DNSGA-II-A | PPS | AE-NSGA-II | KL-NSGA-II | KAEP-NSGA-II |
|---|---|---|---|---|---|---|---|
| DF1 | (5, 10) | 2.6794e-1(2.36e-2)− | 1.3209e-1(1.17e-2)− | 1.0683e-1(4.13e-2)− | 1.2305e-1(8.02e-3)− | 1.6481e-1(1.89e-2)− | 4.8504e-2(7.27e-3) |
|  | (10, 10) | 5.2800e-2(4.76e-3)− | 5.9882e-2(4.88e-3)− | 3.8657e-2(1.47e-2)− | 5.6491e-2(5.05e-3)− | 8.3656e-2(1.11e-2) | 1.4020e-2(1.08e-3) |
|  | (20, 10) | 1.1695e-2(4.38e-4)− | 1.3909e-2(5.39e-4)− | 9.4826e-3(4.12e-4)− | 1.3376e-2(6.86e-4)− | 3.5794e-2(2.43e-3)− | 6.4373e-3(1.88e-4) |
| DF2 | (5, 10) | 2.5814e-1(2.60e-2)− | 8.7572e-2(8.01e-3)+ | 1.3899e-1(1.56e-2)+ | 1.1193e-1(9.92e-3)+ | 1.2842e-1(1.08e-2)+ | 1.8083e-1(2.14e-2) |
|  | (10, 10) | 1.1389e-1(9.98e-3)≈ | 4.4624e-2(6.28e-3)+ | 8.5752e-2(1.28e-2)+ | 8.9802e-2(1.24e-2)+ | 7.5448e-2(6.91e-3)+ | 1.1429e-1(1.24e-2) |
|  | (20, 10) | 4.4869e-2(7.19e-3)+ | 1.7272e-2(3.44e-3)+ | 3.7138e-2(8.08e-3)+ | 3.9524e-2(7.52e-3)+ | 3.7900e-2(6.63e-3)+ | 4.2927e-2(7.53e-3) |
| DF3 | (5, 10) | 4.2387e-1(3.09e-2)− | 4.5158e-1(3.83e-2)− | 8.7014e-1(2.05e-1)− | 4.2211e-1(4.45e-2)− | 6.0592e-1(1.29e-1)− | 1.8831e-1(6.27e-2) |
|  | (10, 10) | 3.5741e-1(3.44e-2)− | 3.7581e-1(3.22e-2)− | 4.3172e-1(6.31e-2)− | 3.5504e-1(3.78e-2)− | 3.9716e-1(3.51e-2)− | 1.8016e-1(3.61e-2) |
|  | (20, 10) | 3.0017e-1(2.64e-2)− | 3.1043e-1(2.29e-2)− | 2.2658e-1(4.72e-2)− | 2.9560e-1(2.38e-2)− | 3.3314e-1(4.02e-2)− | 1.2630e-1(4.90e-2) |
| DF4 | (5, 10) | 1.3812e-1(1.00e-2)− | 1.4629e-1(7.30e-3)− | 7.7441e-1(2.34e-1)− | 1.3395e-1(6.73e-3)− | 1.8548e-1(1.09e-2)− | 1.0577e-1(4.53e-3) |
|  | (10, 10) | 1.0044e-1(4.21e-3)≈ | 1.0015e-1(5.06e-3)− | 1.3039e-1(2.78e-2)− | 9.6082e-2(3.41e-3)≈ | 1.2141e-1(5.20e-3)− | 9.9358e-2(4.17e-3) |
|  | (20, 10) | 8.4014e-2(3.14e-3)+ | 8.4061e-2(2.34e-3)+ | 8.8442e-2(3.53e-3)+ | 8.6463e-2(2.40e-3)+ | 9.5538e-2(4.40e-3)− | 9.0287e-2(2.64e-2) |
| DF5 | (5, 10) | 1.6068e-1(2.04e-2)− | 2.4294e-1(3.42e-2)− | 4.8389e-1(2.15e-1)− | 3.2347e-1(3.28e-2)− | 4.7530e-1(9.29e-2)− | 5.0479e-2(7.87e-3) |
|  | (10, 10) | 4.1844e-2(1.91e-3)≈ | 5.2397e-2(5.43e-2)− | 6.6087e-2(2.62e-2)− | 5.3452e-2(4.11e-3)− | 1.6183e-1(2.98e-2)− | 4.1908e-2(1.17e-3) |
|  | (20, 10) | 1.2360e-2(4.89e-4)− | 1.3147e-2(6.13e-4)− | 2.4710e-2(3.01e-2)− | 1.4661e-2(8.33e-4)− | 3.6919e-2(2.55e-3)− | 7.7099e-3(3.87e-4) |
| DF6 | (5, 10) | 7.4809e+0(7.98e-1)− | 5.7479e+0(3.29e-1)− | 6.0375e+0(4.70e-1)− | 6.4081e+0(4.17e-1)≈ | 6.1602e+0(4.69e-1)− | 3.9864e+0(7.60e-1) |
|  | (10, 10) | 3.1218e+0(5.55e-1)≈ | 2.2170e+0(3.74e-1)≈ | 2.1646e+0(4.19e-1)+ | 2.0956e+0(3.60e-1)+ | 3.7281e+0(3.36e-1)− | 2.2489e+0(6.81e-1) |
|  | (20, 10) | 1.5629e+0(4.27e-1)≈ | 9.8204e-1(2.88e-1)+ | 4.9868e-1(1.49e-1)+ | 9.2374e-1(2.34e-1)+ | 1.1609e+0(1.58e-1)≈ | 1.4194e+0(5.42e-1) |
| DF7 | (5, 10) | 3.1561e-1(7.58e-2)− | 2.8577e-1(5.46e-2)− | 9.3972e-1(5.13e-2)− | 2.2134e-1(3.29e-2)≈ | 1.3055e-1(1.84e-2)− | 2.1868e-1(7.16e-2) |
|  | (10, 10) | 2.8901e-1(7.32e-2)− | 2.6582e-1(5.74e-2)≈ | 3.6192e-1(4.74e-2)− | 1.9632e-1(5.56e-2)≈ | 1.0135e-1(1.55e-2)− | 2.0550e-1(6.94e-2) |
|  | (20, 10) | 2.4868e-1(5.64e-2)− | 2.2043e-1(4.88e-2)− | 3.5063e-1(3.53e-2)− | 1.6456e-1(4.56e-2)≈ | 7.7281e-2(9.26e-3)− | 1.5929e-1(6.77e-2) |
| DF8 | (5, 10) | 1.6536e-2(2.89e-3)≈ | 1.7709e-2(3.26e-3)− | 2.5736e-2(2.00e-2)− | 1.6205e-2(2.29e-3)≈ | 1.1607e-2(5.13e-3)− | 1.4191e-2(1.57e-3) |
|  | (10, 10) | 1.2536e-2(9.49e-3)− | 1.2646e-2(1.06e-3)≈ | 1.2659e-2(9.59e-4)≈ | 1.2083e-2(4.98e-4)≈ | 1.6395e-2(1.17e-2)− | 1.1605e-2(9.08e-4) |
|  | (20, 10) | 1.0876e-2(5.41e-4)− | 1.0865e-2(5.74e-4)− | 1.0385e-2(6.23e-4)− | 1.0654e-2(4.67e-4)− | 4.3624e-2(9.87e-3)− | 9.6912e-3(5.14e-4) |
| DF9 | (5, 10) | 9.6254e-1(1.09e-1)− | 4.8571e-1(6.05e-2)− | 4.3048e-1(7.45e-2)− | 3.4670e-1(2.79e-2)≈ | 5.0570e-1(5.85e-2)− | 1.5494e-1(1.62e-2) |
|  | (10, 10) | 5.5434e-1(5.66e-2)− | 5.2779e-1(3.00e-2)− | 2.4053e-1(5.76e-2)− | 2.1174e-1(5.56e-2)− | 6.8602e-1(3.82e-2)− | 1.0314e-1(1.15e-2) |
|  | (20, 10) | 2.4134e-1(2.77e-2)− | 1.4308e-1(2.36e-2)− | 1.1419e-1(1.09e-2)− | 3.3969e-1(2.61e-2)− | 1.2467e-1(9.61e-3)− | 8.3176e-2(1.64e-2) |
| DF10 | (5, 10) | 9.6302e-2(9.09e-3)+ | 9.5343e-2(6.39e-3)+ | 2.1572e-1(2.83e-2)≈ | 9.7241e-2(6.13e-3)+ | 1.0096e-1(7.41e-3)+ | 1.5617e-1(3.20e-2) |
|  | (10, 10) | 8.7756e-2(5.23e-3)+ | 8.7888e-2(5.66e-3)+ | 1.4855e-1(2.62e-2)≈ | 8.6940e-2(2.36e-3)+ | 9.0937e-2(6.41e-3)+ | 1.2900e-1(2.44e-2) |
|  | (20, 10) | 8.0982e-2(2.70e-3)+ | 8.2106e-2(2.86e-3)+ | 1.1095e-1(8.11e-3)− | 8.3037e-2(3.34e-3)+ | 8.4489e-2(6.09e-3)+ | 1.4261e-1(1.82e-2) |
| DF11 | (5, 10) | 6.4916e-1(2.33e-3)≈ | 6.5498e-1(2.39e-3)≈ | 8.2655e-1(1.29e-1)− | 6.5422e-1(2.04e-3)− | 6.2511e-1(3.28e-3)≈ | 6.4138e-1(2.55e-3) |
|  | (10, 10) | 6.4022e-1(1.62e-3)≈ | 6.4279e-1(1.73e-3)− | 6.6893e-1(1.75e-2)− | 6.4158e-1(1.36e-3)+ | 6.1583e-1(1.94e-3)+ | 6.3353e-1(1.27e-3) |
|  | (20, 10) | 6.3407e-1(7.92e-4)− | 6.3452e-1(8.43e-4)− | 6.4018e-1(2.21e-3)− | 6.3400e-1(7.29e-4)− | 6.0238e-1(3.20e-3)+ | 6.3068e-1(6.38e-4) |
| DF12 | (5, 10) | 2.8745e-1(7.85e-2)− | 3.7847e-1(7.02e-2)− | 4.1140e-1(6.77e-2)− | 2.9393e-1(5.56e-2)− | 5.1923e-1(6.12e-2)− | 3.3026e-1(6.83e-3) |
|  | (10, 10) | 1.4234e-1(6.50e-3)≈ | 1.6006e-1(6.53e-3)− | 2.3039e-1(5.56e-2)− | 1.6302e-1(8.08e-3)− | 3.4003e-1(6.26e-2)− | 1.0526e-1(4.59e-3) |
|  | (20, 10) | 1.0531e-1(2.75e-3)− | 1.1016e-1(3.07e-3)− | 1.3670e-1(4.74e-3)− | 1.1108e-1(3.95e-3)− | 1.5971e-1(1.05e-2)− | 9.1452e-2(2.51e-3) |
| DF13 | (5, 10) | 3.8471e-1(3.37e-2)≈ | 5.3747e-1(1.09e-1)− | 1.1195e+0(4.89e-1)− | 6.5782e-1(7.71e-2)− | 8.7471e-1(1.73e-1)− | 1.6702e-1(7.42e-2) |
|  | (10, 10) | 1.7148e-1(5.39e-2)− | 2.0301e-1(1.20e-2)− | 3.8162e-1(2.89e-1)− | 1.9423e-1(8.56e-2)− | 3.8008e-1(4.78e-2)− | 1.2406e-1(3.41e-3) |
|  | (20, 10) | 1.2380e-1(1.79e-3)− | 1.2553e-1(2.00e-3)− | 1.8880e-1(3.46e-2)− | 1.2335e-1(1.49e-3)− | 1.6332e-1(4.61e-3)− | 1.1008e-1(1.34e-2) |
| DF14 | (5, 10) | 4.8951e-1(9.83e-2)− | 4.8308e-1(1.06e-1)− | 5.4338e-1(9.44e-2)− | 4.9196e-1(8.77e-2)− | 6.1492e-1(8.42e-2)− | 1.9125e-1(2.18e-2) |
|  | (10, 10) | 3.0194e-1(7.25e-2)− | 2.8664e-1(4.33e-2)− | 3.6203e-1(1.22e-1)− | 2.7692e-1(4.85e-2)− | 5.3394e-1(3.84e-2)− | 5.5027e-1(1.93e-2) |
|  | (20, 10) | 2.8636e-1(1.23e-1)− | 9.6683e-1(8.62e-2)− | 2.5533e-1(3.98e-2)− | 7.7764e-1(4.61e-2)− | 2.2787e-1(3.55e-2)− | 1.3434e-1(2.02e-2) |
| +/−/≈ |  | 5/27/10 | 8/30/4 | 5/33/4 | 8/25/9 | 7/29/6 |  |

single-layer denoising autoencoder. This approach learns from the nondominated solutions identified during the dynamic optimization process, resulting in accurate predictions of the moving POS. In addition, the proposed approach can be easily integrated into existing static MOEAs, such as NSGA-II, providing a more solution for DMOPs.

3) KL: KL [46] is a knowledge learning strategy for change response in the dynamic multio-bjective optimization. Unlike other prediction approaches that estimate the future optima from previously obtained solutions, it reacts to changes via learning from the historical search process. KL introduces an autoencoding evolutionary search technique to extract the knowledge within the previous search experience. The extracted knowledge can accelerate convergence as well as introduce diversity for the optimization of the new environment.

## 4.3 Parameter Settings

In this section, the parameter settings of all the compared algorithms are presented. And based on the guidance mentioned in the [50], all of algorithms are implemented on MatlabR2020b under the framework of PlatEMO [51]. Moreover, the parameters of the compared algorithm adopted in the experiment are set

**TABLE II**
The statistics of MHV results ( mean and standard deviation ) achieved by six compared algorithms on bi-objective and tri-objective DF, where '+', '-', and '≈' indicate each compared algorithm is significantly better than, worse than, and tied by KAEP-NSGA-II, respectively.

| Problem | $(\tau_t, n_t)$ | DNSGA-II-B | DNSGA-II-A | PPS | AE-NSGA-II | KL-NSGA-II | KAEP-NSGA-II |
|---|---|---|---|---|---|---|---|
| DF1 | (5, 10) | 2.2337e-1(1.19e-2)− | 3.0887e-1(1.17e-2)− | 3.4351e-1(4.05e-2)− | 3.1967e-1(8.83e-3)− | 2.7919e-1(1.44e-2)− | 4.1734e-1(7.02e-3) |
| | (10, 10) | 4.0487e-1(5.96e-3)− | 3.9464e-1(5.91e-3)− | 4.2589e-1(1.82e-2)− | 3.9883e-1(6.76e-3)− | 3.6586e-1(1.29e-2)− | 4.6098e-1(1.74e-3) |
| | (20, 10) | 4.6440e-1(6.78e-4)− | 4.6100e-1(8.84e-4)− | 4.6798e-1(6.49e-4)− | 4.6172e-1(1.03e-3)− | 4.2855e-1(3.54e-3)− | 4.7276e-1(3.05e-4) |
| DF2 | (5, 10) | 4.2983e-1(2.39e-2)≈ | 6.0666e-1(8.85e-3)+ | 5.4771e-1(1.82e-2)− | 5.8796e-1(8.56e-3)+ | 5.5740e-1(1.14e-2)+ | 5.1251e-1(2.26e-2) |
| | (10, 10) | 5.8831e-1(1.01e-2)− | 6.6976e-1(4.99e-3)+ | 6.1665e-1(1.36e-2)+ | 6.2246e-1(1.27e-2)+ | 6.3039e-1(6.29e-3)+ | 6.0685e-1(1.17e-2) |
| | (20, 10) | 6.7158e-1(7.64e-3)≈ | 7.0516e-1(2.08e-3)+ | 6.7922e-1(9.10e-3)− | 6.8918e-1(4.76e-3)≈ | 6.8173e-1(4.11e-3)≈ | 6.8301e-1(9.58e-3) |
| DF3 | (5, 10) | 1.3090e-1(1.63e-2)− | 1.1797e-1(1.99e-2)− | 4.1982e-2(3.29e-2)− | 1.3102e-1(2.20e-2)− | 7.4265e-2(3.00e-2)− | 2.6424e-1(4.08e-2) |
| | (10, 10) | 1.6850e-1(1.59e-2)− | 1.5911e-1(1.50e-2)− | 1.2812e-1(2.25e-2)− | 1.6860e-1(1.76e-2)− | 1.4404e-1(1.80e-2)− | 2.7248e-1(2.57e-2) |
| | (20, 10) | 1.8685e-1(1.49e-2)− | 1.8099e-1(1.23e-2)− | 1.6933e-1(2.91e-2)− | 1.8964e-1(1.41e-2)− | 1.7942e-1(1.81e-2)− | 3.1131e-1(4.05e-2) |
| DF4 | (5, 10) | 8.0808e-1(4.62e-3)+ | 8.0291e-1(4.43e-3)− | 5.2953e-1(5.84e-2)− | 8.0950e-1(4.13e-3)− | 7.8141e-1(5.58e-3)− | 8.3448e-1(3.01e-3) |
| | (10, 10) | 8.3238e-1(2.12e-3)+ | 8.3313e-1(2.32e-3)− | 8.0705e-1(1.96e-2)− | 8.3643e-1(2.06e-3)− | 8.2504e-1(2.92e-3)− | 8.4769e-1(8.49e-4) |
| | (20, 10) | 8.4765e-1(1.21e-3)+ | 8.4788e-1(9.65e-4)− | 8.4386e-1(8.19e-4)− | 8.4944e-1(5.59e-4)− | 8.4738e-1(2.65e-3)− | 8.5343e-1(3.42e-4) |
| DF5 | (5, 10) | 3.6602e-1(2.14e-2)≈ | 2.9269e-1(2.66e-2)− | 2.3013e-1(6.04e-2)− | 3.0089e-1(3.04e-2)− | 2.2607e-1(1.66e-2)− | 5.1022e-1(1.09e-2) |
| | (10, 10) | 5.2013e-1(2.70e-3)− | 5.0449e-1(5.04e-3)− | 4.6329e-1(4.19e-2)− | 5.0273e-1(5.90e-3)− | 3.7831e-1(2.54e-2)− | 5.6126e-1(1.78e-3) |
| | (20, 10) | 5.6498e-1(7.04e-4)− | 5.6378e-1(8.97e-4)− | 5.5115e-1(3.32e-2)− | 5.6152e-1(1.29e-3)− | 5.2926e-1(3.82e-3)− | 5.7216e-1(6.17e-4) |
| DF6 | (5, 10) | 1.0198e-1(3.06e-2)− | 1.2336e-1(2.04e-2)− | 1.0406e-2(2.70e-3)− | 1.6332e-1(2.58e-2)≈ | 4.9682e-2(7.95e-3)− | 3.0465e-1(8.17e-2) |
| | (10, 10) | 1.9775e-1(6.67e-2)− | 2.0586e-1(5.97e-2)− | 1.5705e-1(6.30e-2)− | 2.5993e-1(3.89e-2)− | 1.1761e-1(3.42e-2)− | 3.0766e-1(1.06e-1) |
| | (20, 10) | 2.8931e-1(7.41e-2)≈ | 3.6524e-1(8.42e-2)+ | 3.9343e-1(7.65e-2)+ | 3.3602e-1(9.90e-2)+ | 2.9097e-1(5.63e-2)≈ | 2.8414e-1(1.03e-1) |
| DF7 | (5, 10) | 3.5015e-1(2.63e-2)− | 3.6157e-1(1.22e-2)− | 3.2526e-1(2.44e-2)− | 3.9490e-1(1.06e-2)≈ | 4.0731e-1(1.55e-2)≈ | 4.1032e-1(1.33e-2) |
| | (10, 10) | 3.6523e-1(2.50e-2)− | 3.7327e-1(1.51e-2)− | 3.4052e-1(2.15e-2)− | 4.0429e-1(1.38e-2)− | 4.2381e-1(5.78e-3)+ | 4.1525e-1(1.25e-2) |
| | (20, 10) | 3.9141e-1(1.54e-2)− | 3.9720e-1(1.57e-2)− | 3.6548e-1(2.53e-2)− | 4.1740e-1(7.86e-3)− | 4.3494e-1(4.13e-3)+ | 4.2676e-1(1.11e-2) |
| DF8 | (5, 10) | 6.2200e-1(2.20e-3)≈ | 6.2055e-1(3.10e-3)≈ | 5.9388e-1(3.94e-2)− | 6.2402e-1(5.03e-4)≈ | 6.5997e-1(2.85e-3)≈ | 6.6347e-1(9.34e-4) |
| | (10, 10) | 6.2429e-1(1.09e-3)− | 6.2261e-1(2.42e-3)− | 6.2059e-1(2.55e-3)− | 6.2488e-1(1.25e-4)≈ | 6.5967e-1(1.27e-2)+ | 6.2491e-1(1.30e-4) |
| | (20, 10) | 6.2493e-1(1.13e-3)− | 6.2455e-1(1.29e-3)− | 6.2350e-1(1.58e-3)− | 6.2537e-1(6.35e-5)− | 6.5753e-1(1.48e-2)+ | 6.2546e-1(6.75e-5) |
| DF9 | (5, 10) | 1.5548e-1(2.02e-2)− | 1.6105e-1(2.01e-2)− | 1.9502e-1(2.55e-2)− | 2.0633e-1(1.66e-2)≈ | 1.4208e-1(1.63e-2)− | 3.6595e-1(1.48e-2) |
| | (10, 10) | 2.8839e-1(1.05e-2)− | 3.0545e-1(1.91e-2)− | 3.1004e-1(2.56e-2)− | 3.2826e-1(1.91e-2)− | 2.6237e-1(1.94e-2)− | 4.2254e-1(1.26e-2) |
| | (20, 10) | 3.8891e-1( 8.65e-3)− | 4.1350e-1(1.04e-2)− | 4.1600e-1(1.14e-2)− | 4.1507e-1(1.49e-2)− | 4.0040e-1(1.03e-2)− | 4.4914e-1(1.87e-2) |
| DF10 | (5, 10) | 6.9082e-1(5.88e-3)≈ | 6.9122e-1(5.04e-3)≈ | 5.1615e-1(2.54e-2)− | 6.9259e-1(3.05e-3)+ | 6.9076e-1(4.28e-3)≈ | 6.7890e-1(1.03e-2) |
| | (10, 10) | 7.0018e-1(3.51e-3)+ | 6.9972e-1(2.36e-3)+ | 6.0178e-1(2.73e-2)− | 7.0184e-1(2.03e-3)+ | 6.9896e-1(4.13e-3)+ | 6.9226e-1(5.29e-3) |
| | (20, 10) | 7.0491e-1(2.07e-3)+ | 7.0381e-1(2.24e-3)+ | 6.6378e-1(1.42e-2)− | 7.0625e-1(1.49e-3)+ | 7.0482e-1(2.08e-3)+ | 7.0045e-1(4.43e-3) |
| DF11 | (5, 10) | 8.5757e-2(7.47e-4)+ | 8.5186e-2(7.50e-4)− | 5.5396e-2(1.44e-2)− | 8.6805e-2(6.15e-4)≈ | 1.0222e-1(2.70e-4)+ | 8.7313e-2(7.14e-4) |
| | (10, 10) | 8.7026e-2(5.78e-4)+ | 8.6635e-2(7.88e-4)− | 8.2088e-2(4.61e-3)− | 8.7534e-2(6.37e-4)≈ | 1.0342e-1(1.02e-3)+ | 8.7693e-2(5.99e-4) |
| | (20, 10) | 8.7735e-2(6.68e-4)≈ | 8.7630e-2(4.79e-4)− | 8.7138e-2(5.95e-4)− | 8.7786e-2(4.99e-4)− | 1.0315e-1( 3.80e-3)+ | 8.7981e-2(5.05e-4) |
| DF12 | (5, 10) | 5.7803e-1(2.09e-2)− | 5.5053e-1(1.28e-2)− | 4.0293e-1(2.82e-2)− | 5.4329e-1(1.33e-2)− | 5.3195e-1(1.55e-2)− | 6.5934e-1(3.62e-3) |
| | (10, 10) | 6.4947e-1(5.46e-3)− | 6.3895e-1(5.72e-3)− | 4.3313e-1(3.29e-2)− | 6.4404e-1(3.78e-3)− | 5.7614e-1(1.84e-2)− | 6.7507e-1(2.61e-3) |
| | (20, 10) | 6.7169e-1(1.96e-3)− | 6.6830e-1(3.27e-3)− | 2.6996e-1(4.53e-3)− | 6.7290e-1(1.60e-3)− | 6.4734e-1(6.23e-3)− | 6.8319e-1(1.75e-3) |
| DF13 | (5, 10) | 4.2036e-1(1.88e-2)− | 3.5126e-1(4.37e-2)− | 2.5013e-1(5.11e-2)− | 2.9594e-1(2.72e-2)− | 2.5278e-1(3.09e-2)− | 5.7465e-1(7.15e-3) |
| | (10, 10) | 5.6610e-1(5.31e-3)− | 3.3936e-1(8.93e-3)− | 4.8737e-1(8.68e-2)− | 5.4723e-1(7.74e-3)− | 4.2814e-1(2.66e-2)− | 6.1508e-1(3.81e-3) |
| | (20, 10) | 6.1371e-1(1.81e-3)− | 6.1151e-1(2.05e-3)− | 6.2009e-1(4.04e-3)− | 6.1346e-1(1.64e-3)− | 5.7821e-1(4.54e-2)− | 6.3273e-1(1.53e-3) |
| DF14 | (5, 10) | 1.2717e-1(2.29e-2)− | 1.1116e-1(5.95e-2)− | 1.2346e-1(2.53e-2)− | 9.8418e-2(3.91e-2)− | 5.2131e-2(1.73e-2)− | 4.4793e-1(2.48e-2) |
| | (10, 10) | 2.5819e-1(5.21e-2)− | 2.4636e-1(5.15e-2)− | 2.1598e-1(5.20e-2)− | 2.4138e-1(4.11e-2)− | 1.6153e-1(3.69e-2)− | 3.9274e-1(2.88e-2) |
| | (20, 10) | 4.1603e-1(2.53e-2)− | 4.1067e-1(1.99e-2)− | 3.3358e-1(2.42e-2)− | 4.1243e-1(2.86e-2)− | 3.0493e-1(4.10e-2)− | 4.4919e-1(1.61e-2) |
| +/−/≈ | | 2/33/7 | 6/34/2 | 2/39/1 | 6/27/9 | 10/26/6 | |

according to the their original papers. Other common parameters are briefly presented as follows.

1) Population Size: The population size N is set to 100 for all the DF problems.

2) Number of Detectors: The number of detectors is set to 10% of the population size same as the setting of most DMOEAs. Specifically, in this paper, ten solutions are randomly selected and re-evaluated to make sure whether the environment has changed in each generations of all the compared algorithms.

3) Reproduction Operators: The baseline optimizer NSGAII use the simulated binary crossover (SBX) [52] and the polynomial mutation (PM) [53] as the reproduction operators. Specifically, the crossover probability pc and distribution index ηc in SBX are set to 0.9 and 20, respectively. The distribution index ηm is set to 20 and the mutation probability pm is set to 1/n in PM, where n is the number of decision variables.

4) Termination Condition: Considering that the environmental changes are related to the number of generations given in (14), this article uses the maximum number of generations as the termination condition for each test function. At the beginning of the algorithm, the population iterates for 100 generations, which enables the population to converge [54]. Moreover, as suggested in [29], the number of environmental changes is fixed to 20 in each run,

TABLE III
The statistics of MGD results ( mean and standard deviation ) achieved by six compared algorithms on bi-objective and tri-objective DF, where '+', '-', and '≈' indicate each compared algorithm is significantly better than, worse than, and tied by KAEP-NSGA-II, respectively.

| Problem | (τ, n) | DNSGA-II-B | DNSGA-II-A | PPS | AE-NSGA-II | KL-NSGA-II | KAEP-NSGA-II |
|---|---|---|---|---|---|---|---|
| DF1 | (5, 10) | 2.4088e-2(8.02e-3)− | 2.8999e-2(3.12e-3)− | 2.2573e-1(1.05e-2)− | 2.5817e-2(2.22e-3)− | 4.1586e-2(4.92e-3)− | 8.6582e-3(1.52e-3) |
| | (10, 10) | 8.5496e-3(1.06e-3)− | 9.9974e-3(1.08e-3)− | 6.1328e-3(3.14e-3)− | 9.2532e-3(1.14e-3)− | 1.6867e-2(2.83e-3)− | 1.5622e-3(1.79e-4) |
| | (20, 10) | 1.0830e-3(5.12e-5)− | 1.3543e-3(7.56e-5)− | 7.9323e-4(5.62e-5)− | 1.2872e-3(9.17e-5)− | 5.1469e-3(4.93e-4)− | 4.0739e-4(2.32e-5) |
| DF2 | (5, 10) | 4.6137e-2(7.23e-3)− | 1.6067e-2(2.91e-3)− | 1.7385e-2(4.74e-3)− | 1.5419e-2(1.79e-3)≈ | 3.4789e-2(7.96e-3)− | 1.0884e-2(2.31e-3) |
| | (10, 10) | 4.6857e-3(6.58e-4)− | 4.0860e-3(4.33e-4)− | 4.8199e-3(1.37e-3)− | 5.5078e-3(5.88e-4)− | 9.4609e-3(9.02e-4)− | 1.4456e-3(1.74e-4) |
| | (20, 10) | 6.7236e-4(4.11e-5)− | 8.7719e-4(3.58e-4)− | 9.0220e-4(7.48e-5)− | 8.7310e-4(4.95e-5)− | 4.2259e-3(1.55e-4)− | 3.2427e-4(2.08e-5) |
| DF3 | (5, 10) | 4.1747e-2(7.86e-3)− | 5.4697e-2(1.25e-2)− | 3.1514e-1(1.16e-1)− | 4.9226e-2(1.30e-2)− | 1.2866e-1(7.29e-2)− | 9.2625e-3(3.78e-3) |
| | (10, 10) | 1.1220e-2(1.47e-3)− | 1.2536e-2(1.59e-3)− | 1.9803e-2(2.29e-2)− | 1.2880e-2(1.86e-3)− | 2.9387e-2(7.83e-3)− | 3.4261e-3(1.19e-3) |
| | (20, 10) | 3.1474e-3(2.96e-4)− | 3.1532e-3(3.09e-4)− | 1.9778e-3(9.80e-4)≈ | 2.9104e-3(2.98e-4)− | 6.4790e-3(9.24e-4)− | 2.0114e-3(7.70e-4) |
| DF4 | (5, 10) | 2.8197e-2(5.72e-3)− | 3.0771e-2(6.74e-3)− | 5.7502e-1(1.44e-1)− | 2.5097e-2(4.34e-3)≈ | 9.2380e-2(3.33e-2)− | 1.5695e-2(3.24e-3) |
| | (10, 10) | 1.0709e-2(2.33e-3)− | 1.2836e-2(3.27e-3)− | 5.4818e-2(3.08e-2)− | 1.2501e-2(2.73e-3)− | 3.2249e-2(7.58e-3)− | 9.0798e-3(2.28e-3) |
| | (20, 10) | 7.0782e-3(1.87e-3)≈ | 8.1208e-3(2.17e-3)− | 8.0092e-3(9.57e-4)− | 6.8681e-3(1.67e-3)− | 1.9164e-2(1.37e-3)− | 6.0741e-3(1.50e-3) |
| DF5 | (5, 10) | 5.5289e-2(6.86e-3)≈ | 9.2024e-2(1.93e-2)− | 1.7921e-1(7.13e-2)− | 2.0251e-1(5.28e-2)− | 2.9180e-1(5.58e-2)− | 3.9844e-2(9.91e-3) |
| | (10, 10) | 2.0756e-2(2.00e-3)≈ | 2.6002e-2(2.60e-3)≈ | 4.4207e-2(2.20e-2)− | 1.2206e-1(5.59e-2)− | 2.5844e-1(5.88e-2)− | 2.0076e-2(4.66e-3) |
| | (20, 10) | 1.3428e-2(3.60e-3)+ | 2.4199e-2(2.26e-2)≈ | 2.8973e-2(2.74e-2)≈ | 9.8532e-2(3.12e-2)− | 2.9736e-1(3.44e-2)− | 2.5970e-2(7.45e-3) |
| DF6 | (5, 10) | 6.1166e+0(4.64e-1)− | 5.0792e+0(3.45e-1)− | 7.5899e+0(6.64e-1)− | 4.2524e+0(3.04e-1)≈ | 7.4654e+0(3.99e-1)− | 2.6999e+0(3.79e-1) |
| | (10, 10) | 2.1701e+0(2.94e-1)− | 1.8457e+0(2.12e-1)− | 2.0783e+0(3.56e-1)− | 1.6028e+0(1.47e-1)≈ | 4.7262e+0(3.37e-1)− | 2.2568e+0(2.55e-1) |
| | (20, 10) | 7.0684e-1(1.39e-1)− | 4.4139e-1(8.83e-2)+ | 3.3408e-1(8.50e-2)+ | 8.3297e-1(7.83e-2)+ | 1.9975e+0(2.37e-1)− | 5.2233e-1(1.47e-1) |
| DF7 | (5, 10) | 3.4802e-2(2.51e-2)− | 4.8381e-2(2.20e-2)− | 6.6848e-2(4.13e-2)− | 1.6547e-2(8.81e-3)− | 1.8991e-2(4.31e-3)− | 3.3116e-3(1.19e-3) |
| | (10, 10) | 2.0066e-2(1.22e-2)− | 2.4038e-2(9.47e-3)− | 4.2594e-2(3.25e-2)− | 1.0962e-2(6.99e-3)− | 9.8633e-3(1.79e-3)− | 2.6148e-3(9.11e-4) |
| | (20, 10) | 7.7517e-3(3.33e-3)− | 9.0569e-3(4.67e-3)− | 3.6394e-3(1.79e-3)− | 4.2505e-3(1.48e-3)− | 4.9706e-3(9.23e-4)− | 1.9303e-3(4.53e-4) |
| DF8 | (5, 10) | 9.1886e-4(2.21e-4)+ | 1.5457e-3(6.86e-4)≈ | 4.9087e-3(8.64e-3)≈ | 7.0869e-4(1.04e-4)+ | 5.7653e-3(1.07e-3)− | 1.7377e-3(5.77e-4) |
| | (10, 10) | 7.2514e-4(7.30e-5)+ | 9.6616e-4(5.00e-4)≈ | 1.3833e-3(5.71e-4)− | 6.9299e-4(8.15e-6)+ | 5.5155e-3(1.59e-3)− | 8.3208e-4(1.34e-4) |
| | (20, 10) | 7.1484e-4(1.36e-4)≈ | 6.8725e-4(2.91e-5)≈ | 6.9384e-4(5.72e-5)≈ | 6.7148e-4(7.69e-6)+ | 5.5562e-3(1.50e-3)− | 6.7712e-4(1.06e-5) |
| DF9 | (5, 10) | 7.0301e-1(5.80e-2)− | 3.8326e-1(4.32e-2)− | 2.9156e-1(4.09e-2)− | 2.4020e-1(4.73e-2)− | 4.0045e-1(8.02e-2)− | 6.8976e-2(1.67e-2) |
| | (10, 10) | 3.5898e-1(3.82e-2)− | 1.9143e-1(4.01e-2)− | 1.4689e-1(2.91e-2)− | 1.2449e-1(4.45e-2)− | 1.4581e-1(2.76e-2)− | 2.9093e-2(6.00e-3) |
| | (20, 10) | 1.4350e-1(2.14e-2)− | 8.5517e-2(2.36e-2)− | 3.3127e-2(6.74e-3)− | 7.9094e-2(2.81e-2)− | 1.5336e-2(8.20e-3)− | 1.5300e-2(2.45e-3) |
| DF10 | (5, 10) | 9.8571e-3(1.13e-3)+ | 1.0997e-2(3.23e-3)≈ | 9.6570e-2(1.54e-2)− | 8.3921e-3(1.44e-3)+ | 1.3483e-2(4.65e-3)+ | 1.4581e-2(2.05e-3) |
| | (10, 10) | 7.7540e-3(1.26e-3)+ | 8.5638e-3(2.37e-3)≈ | 5.5575e-2(1.36e-2)− | 7.0069e-3(1.23e-3)+ | 1.0177e-2(2.36e-3)+ | 9.1592e-3(1.90e-3) |
| | (20, 10) | 3.7096e-3(1.59e-3)≈ | 8.2055e-3(1.64e-3)≈ | 1.4731e-2(6.70e-3)− | 1.8638e-3(1.08e-3)+ | 8.2418e-3(1.41e-3)≈ | 7.2717e-3(1.63e-3) |
| DF11 | (5, 10) | 6.7106e-2(4.23e-4)− | 6.7719e-2(6.69e-4)− | 1.3001e-1(6.68e-2)− | 6.9073e-2(6.71e-4)− | 6.5240e-2(8.12e-4)≈ | 6.5131e-2(5.14e-4) |
| | (10, 10) | 6.5104e-2(3.36e-4)− | 6.5711e-2(3.52e-4)− | 2.2802e-2(6.67e-3)− | 6.5626e-2(2.79e-4)− | 6.4319e-2(6.31e-4)+ | 6.3704e-2(1.97e-4) |
| | (20, 10) | 6.3791e-2(1.48e-4)− | 6.3967e-2(2.03e-4)− | 6.5049e-2(3.93e-4)− | 6.3874e-2(1.69e-4)− | 6.2050e-2(4.19e-4)+ | 6.3204e-2(8.18e-5) |
| DF12 | (5, 10) | 3.8600e-2(5.23e-3)− | 4.9984e-2(1.24e-2)− | 1.2968e-1(2.79e-2)− | 5.2229e-2(5.54e-3)− | 6.1238e-2(2.14e-2)− | 1.5905e-2(3.61e-3) |
| | (10, 10) | 1.8583e-2(1.80e-2)− | 2.3380e-2(2.29e-3)− | 6.1237e-2(1.94e-2)− | 2.2040e-2(5.53e-3)− | 3.2610e-2(8.79e-3)− | 1.4745e-2(2.91e-3) |
| | (20, 10) | 1.2735e-2(1.39e-3)− | 1.3600e-2(1.70e-3)− | 2.0902e-2(1.62e-3)− | 1.3412e-2(1.86e-3)− | 2.1830e-2(2.89e-3)− | 1.0898e-2(2.07e-3) |
| DF13 | (5, 10) | 7.5856e-2(8.18e-3)≈ | 1.1352e-1(2.27e-2)− | 3.0067e-1(1.49e-1)− | 1.3000e-1(1.65e-2)− | 2.3777e-1(4.24e-2)− | 1.9259e-2(1.91e-3) |
| | (10, 10) | 1.9792e-2(1.39e-3)− | 2.6103e-2(2.39e-3)− | 7.5117e-2(7.11e-2)− | 2.4030e-2(1.85e-3)− | 7.3350e-2(1.34e-2)− | 1.0450e-2(2.85e-3) |
| | (20, 10) | 1.0397e-2(2.68e-4)− | 1.0910e-2(3.20e-4)− | 5.9892e-3(6.98e-4)− | 1.0374e-2(3.14e-4)− | 1.9741e-2(2.82e-3)− | 7.7312e-3(3.20e-4) |
| DF14 | (5, 10) | 8.6354e-1(8.40e-2)− | 8.4313e-1(8.23e-2)− | 3.5769e-1(2.89e-2)≈ | 8.1365e-1(1.39e-1)− | 9.4544e-1(1.07e-1)− | 3.1417e-1(5.13e-2) |
| | (10, 10) | 9.6896e-1(8.49e-2)− | 9.1496e-1(6.55e-2)− | 3.6366e-1(5.97e-2)≈ | 8.9541e-1(7.42e-2)− | 1.0747e+0(8.25e-2)− | 4.0518e-1(5.81e-2) |
| | (20, 10) | 9.3632e-1(6.50e-3)− | 9.5419e-1(7.85e-3)− | 4.8228e-1(4.11e-2)≈ | 9.6381e-1(7.44e-2)− | 1.1177e+0(6.43e-2)− | 4.4276e-1(6.62e-2) |
| +/−/≈ | | 5/31/6 | 1/34/7 | 1/34/7 | 3/32/7 | 1/37/4 | |

accordingly the maximum number of generation is set to set to $20 \times \tau_1 + 100$.

## 4.4 Performance Metrics

In this section, four widely used performance metrics are introduced, which is adopted to make statistically robust comparisons in the experimental studies. A simple introduction of these metrics is presented as follows.

1) Mean Inverted Generational Distance (MIGD): The MIGD metric calculates the average values of inverted generational distance (IGD) [55] of all time steps over a run. A smaller value of MIGD indicates a better performance of an algorithm in terms of convergence and the diversity. Specifically, a set of around 15000 points , uniformly distributed in the true DPOF, is denoted as Pt. Meanwhile, let $P_t^*$ represent an approximation of the DPOF at time t, the MIGD can be calculated as follows:

$$MIGD = \frac{1}{T} \sum_{i=1}^{T} IGD(P_t^*, P_t) \qquad (15)$$

where T is the number of environmental changes.

2) Mean Hypervolume (MHV): The MHV metric calculates the average values of Hypervolume (HV) [56]. Same as MIGD, MHV metric is able to give a comprehensive information, including the convergence and diversity, simultaneously. The larger is the MHV value, the better is the quality of solutions obtained an algorithm. The MHV can be expressed by

$$MHV = \frac{1}{T}\sum_{i=1}^{T} HV(P_t^*) \qquad (16)$$

where HV(S) is the hypervolume of a set of solutions. To compute the hypervolume of $P_t^*$, a reference point $(z_1 + 0.1, \cdots, z_m + 0.1)$ is used, where $z_j$ is the maximum value of the $j$-th objective of the true DPOF at time t and M is the number of objectives.

3) Mean Generational Distance (MGD): The MGD caculates the average vales of generational distance (GD) [57], which measures the convergence of the obtained solutions towards the true DPOF of each time step. The expression formula of MGD can be given as follows

$$MGD = \frac{1}{T}\sum_{i=1}^{T} GD(P_t^*) \qquad (17)$$

4) Mean Schott's Spacing Metric (MSP): The MSP based on the Schott's spacing (SP) metric [58] is widely used to measure the diversity of the obtained solutions by estimating the distribution degree of the discovered Pareto front [59]. The expression formula of MSP can be given as follows

$$MSP = \frac{1}{T}\sum_{i=1}^{T} SP(P_t^*) \qquad (18)$$

## 5 Experimental Results and Analysis

In this section, we present the results of the proposed KAEP-NSGA-II and the compared algorithms on DMOPs,including the mean and standard deviations of four metric values. Each algorithm is run for 20 times on each DMOP independently, where the Friedman test [60] is conducted on the experimental results. "+", "−" and "≈" denote that the proposed approach is statistically significantly worse, better, and similar to the compared dynamic multi-objective method which shares the same multi-objective optimizer, respectively.

### 5.1 Experimental Results on DMOPs

The statistical results have been organized into Tables I-IV. Within these tables, the best values achieved by one of the six algorithms are indicated in bold face. As observed from Table I and Table II, the proposed KAEP-NSGA-II obtains the better MIGD and MHV values against the compared algorithms on most cases under different dynamic configurations. for solving DF test problems. Specifically, the proposed KAEP-NSGA-II achieves 27 out of 42 best MIGD values and 28 out of 42 best MHV values, respectively.

As shown in Table I, KAEP-NSGA-II performs the best on the majority of the test problems and mainly loses on DF2, DF7, DF9-10 in terms of the MIGD metric. DNSGA-II-A has performed the best on DF2 with different frequency of environmental changes. To some extent, this is because DF2 has a simple dynamic on the PS, and its PF remains stationary over time. In situations where environmental changes are not very complex, reevaluating the optimal solution from the previous environment as initialization is comparatively effective. while for the AE-NSGA-II, KL-NSGA-II and the proposed KAEP-NSGA-II, the knowledge transfer DMOEAs based on the auto-encoding evolutionary search, did not perform well on this type of problem, which may be due to negative transfer. Meanwhile, it worth noting that as $\tau_t$ gets larger, the performance of DNSGA-II-A and DNSGA-II-B gets better, and in some testing problems, it even performs the best. Such as DNSGA-II-B achives the best performance on DF4 and DF10 with time related parameter $\tau_t = 20$. For the KL-NSGAII, it performs best than the other compared algorithms on the DF7 and DF11. Especially, the DPOS of DF7 is dynamic, but its centroid remains unchanged which lead to be invalidation for

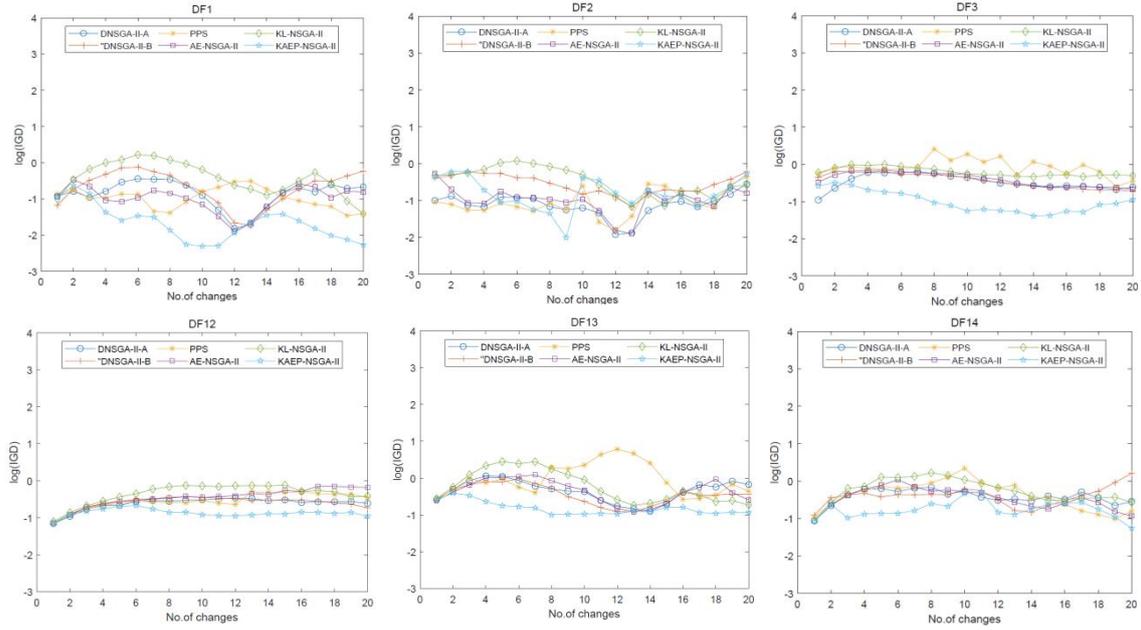

Fig.3 log(IGD) values obtained by six algorithms for different test functions with $\tau_t = 5$ and $n_t = 10$

centroid-based prediction methods in some degree. So the PPS strategy and the proposed KAEP strategy that rely primarily on the centroid to initialize in a new environment are ineffective. Furthermore, Table II list the MHV metric values for the DF test problems. and the experimental results show that the proposed KAEP-NSGA-II consistently outperforms competing algorithms for most test problems. In addition to Table I and Table II, The log(IGD) values obtained by six algorithms after each change for certain problems are shown in Fig.3 that provide a more clear comparsion. Among these algorithms, KAEP-NSGA-II consistently exhibits favorable results in terms of convergence and diversity maintenance in most cases.

As indicated in Table III and Table IV, the KAEP-NSGA-II has the best performance in most cases from the point of view of convergence and distribution of solutions. Specifically, the proposed KAEP-NSGA-II achieves 32 out of 42 best MGD values and 28 out of 42 best MSP values, respectively. With respect to the MGD, KAEP-NSGA-II significaantly outperforms its competitors at all dynamic parameter configurations on DF1, DF2, DF4,

DF7, DF9 and DF12-13, which indicates that the proposed prediction strategy can respond to the environmental changes more efficiently in the most of the scenarios when equipped with the same baseline optimizer. On the other hand, it can be observed from Table IV that KAEP-NSGA-II obtains the best results on the majority of the DF test problems, implying that it maintains better distribution of its approximations over changes than the other compared algorithms in most cases. However, it performs slightly worse than DNSGA-II for DF5, DF8, DF11 with different parameter $\tau_t$. In addition, AE-NSGA-II has obtained best performance to maintain uniformity of solutions on the DF10. PPS-NSGA-II performs better than KAEP-NSGA-II for DF3 and DF8 with a slower changes (i.e., $\tau\tau = 20$). For all the test problems, KL-NSGA-II fail to show encouraging performance from the results showed in Table IV.

In summary, KAEP performs quite well compared with other centroid-based (PPS) and autoencoding search-based (AE and KL) methods on most cases of DF test problems. In particular, KAEP performs significantly best than all other compared algorithms on DF4, DF9, and DF12-14 with irregular change.

| Problem | $(\tau, \mathbf{n})$ | DNSGA-II-B | DNSGA-II-A | PPS | AE-NSGA-II | KL-NSGA-II | KAEP-NSGA-II |
|---|---|---|---|---|---|---|---|
| DF1 | (5, 10) | 5.2587e-2(7.28e-3)− | 3.7399e-2(3.50e-3)− | 2.9356e-2(6.42e-3)− | 3.4650e-2(4.04e-3)− | 5.6072e-2(6.91e-3)− | 1.5248e-2(1.77e-3) |
|  | (10, 10) | 1.9425e-2(1.62e-3)− | 2.0117e-2(1.27e-3)− | 1.5259e-2(3.89e-3)− | 2.0481e-2(1.94e-3)− | 3.1782e-2(3.73e-3)− | 8.3136e-3(3.91e-4) |
|  | (20, 10) | 7.8669e-3(4.17e-4)− | 8.5112e-3(4.69e-4)− | 7.6028e-3(3.71e-4)− | 8.4429e-3(3.91e-4)− | 1.6159e-2(1.11e-3)− | 6.9172e-3(1.56e-4) |
| DF2 | (5, 10) | 4.4079e-2(8.24e-3)− | 3.3053e-2(9.02e-3)− | 2.9985e-2(5.42e-3)− | 3.3511e-2(6.38e-3)− | 6.5501e-2(1.73e-2)− | 1.9770e-2(4.93e-3) |
|  | (10, 10) | 1.3700e-2(1.70e-3)− | 1.3202e-2(9.57e-4)− | 1.3981e-2(1.66e-3)− | 1.4948e-2(1.55e-3)− | 2.3239e-2(2.83e-3)− | 8.8029e-3(1.15e-3) |
|  | (20, 10) | 7.0945e-3(9.25e-4)≈ | 6.8046e-3(2.68e-4)≈ | 6.6810e-3(1.32e-3)− | 7.1894e-3(5.58e-4)− | 1.0747e-2(4.00e-4)− | 6.5934e-3(9.39e-4) |
| DF3 | (5, 10) | 3.0828e-2(0.00e+0)≈ | 2.8727e-2(2.39e-3)≈ | 5.9897e-2(1.48e-2)− | 2.9581e-2(3.18e-3)≈ | 3.4709e-2(6.92e-3)≈ | 3.4509e-2(1.22e-2) |
|  | (10, 10) | 1.8046e-2(3.37e-3)≈ | 1.7426e-2(3.6e-3)≈ | 2.0384e-2(4.77e-3)− | 1.8646e-2(5.32e-3)≈ | 2.8991e-2(1.41e-2)− | 1.7389e-2(4.43e-3) |
|  | (20, 10) | 1.2607e-2(2.08e-3)≈ | 1.2885e-2(1.78e-3)≈ | 1.0414e-2(4.00e-3)+ | 1.2189e-2(2.34e-3)≈ | 2.1374e-2(5.16e-3)− | 1.5423e-2(6.36e-3) |
| DF4 | (5, 10) | 1.2300e-1(2.51e-2)− | 1.3342e-1(4.24e-2)− | 9.3412e-1(8.17e-2)− | 1.1027e-1(1.85e-2)− | 2.6189e-1(1.10e-1)− | 9.2507e-2(2.53e-3) |
|  | (10, 10) | 6.3009e-2(1.04e-2)≈ | 7.8951e-2(2.82e-2)− | 1.1384e-1(4.27e-2)− | 7.1142e-2(1.53e-2)− | 1.4829e-1(4.36e-2)− | 5.7215e-2(1.90e-2) |
|  | (20, 10) | 4.7417e-2(1.41e-2)≈ | 5.5460e-2(2.15.85e-2)− | 4.3375e-2(6.53e-3)≈ | 4.3932e-2(1.15.2e-2)≈ | 1.1120e-1(8.00e-2)− | 3.8140e-2(1.14e-2) |
| DF5 | (5, 10) | 5.2217e-2(6.63e-3)− | 6.7153e-2(8.63e-3)− | 1.1677e-1(3.23e-2)− | 7.3374e-2(1.02e-2)− | 1.3955e-1(2.46e-2)− | 2.5861e-2(4.81e-3) |
|  | (10, 10) | 2.0389e-2(1.25e-3)− | 2.4576e-2(3.50e-3)− | 3.2281e-2(9.63e-3)− | 3.3849e-2(6.62e-3)− | 7.6302e-2(1.44e-2)− | 1.2617e-2(1.45e-3) |
|  | (20, 10) | 1.0650e-2(1.32e-3)+ | 1.1845e-2(4.39e-3)+ | 1.4800e-2(1.03e-2)≈ | 2.5447e-2(4.93e-3)− | 6.3359e-2(6.87e-3)− | 1.2490e-2(1.78e-3) |
| DF6 | (5, 10) | 3.7766e+0(7.21e-1)− | 3.0574e+0(4.26e-1)− | 4.8117e+0(7.64e-1)− | 2.7835e+0(3.04e-1)− | 6.0057e+0(7.76e-1)− | 1.5724e+0(3.88e-1) |
|  | (10, 10) | 1.4152e+0(2.90e-1)− | 1.2817e+0(2.70e-1)− | 1.4227e+0(3.47e-1)− | 1.0573e+0(1.23e-1)− | 5.1866e+0(5.95e-1)− | 7.4076e-1(2.51e-1) |
|  | (20, 10) | 4.0743e-1(1.02e-1)− | 3.0315e-1(6.37e-2)≈ | 3.0822e-1(4.87e-2)− | 3.0393e-1(6.96e-2)≈ | 5.3177e+0(4.14e-1)− | 2.6046e-1(1.45e-1) |
| DF7 | (5, 10) | 7.7325e-2(7.30e-2)− | 9.7641e-2(3.74e-2)− | 2.1336e-1(6.67e-2)− | 6.2820e-2(2.28e-2)− | 2.7709e-2(1.21e-2)− | 4.1757e-2(2.05e-2) |
|  | (10, 10) | 5.9483e-2(3.01e-2)− | 6.6022e-2(2.1.97e-2)− | 6.3918e-2(4.02e-2)− | 4.9004e-2(2.64e-2)≈ | 4.6387e-2(6.95e-3)− | 3.4648e-2(1.35e-2) |
|  | (20, 10) | 4.3178e-2(7.05e-3)− | 4.7425e-2(2.25e-2)− | 8.3136e-2(1.02e-2)− | 2.7024e-2(5.05e-3)− | 3.3230e-2(6.28e-3)− | 2.6201e-2(1.10e-2) |
| DF8 | (5, 10) | 2.4384e-2(4.33e-3)≈ | 2.6062e-2(5.52e-3)− | 3.0705e-2(1.45e-2)≈ | 2.5693e-2(2.71e-3)≈ | 3.1326e-2(7.51e-3)≈ | 2.7139e-2(2.47e-3) |
|  | (10, 10) | 2.1231e-2(1.52e-3)≈ | 2.0399e-2(2.3.21e-3)≈ | 2.0789e-2(5.44e-3)≈ | 2.2046e-2(1.91e-3)− | 2.2236e-2(3.39e-3)≈ | 2.0644e-2(1.73e-3) |
|  | (20, 10) | 2.0314e-2(1.67e-3)− | 1.8967e-2(1.96e-3)− | 1.7091e-2(2.25e-3)+ | 2.0214e-2(1.91e-3)− | 1.9273e-2(2.93e-3)− | 1.7846e-2(1.24e-3) |
| DF9 | (5, 10) | 5.8338e-1(9.85e-2)− | 4.0861e-1(9.13e-2)− | 3.1453e-1(2.84e-2)− | 2.6611e-1(4.13e-2)− | 4.5624e-1(9.59e-2)− | 1.6495e-1(2.29e-2) |
|  | (10, 10) | 3.0023e-1(8.86e-2)− | 2.2727e-1(4.86e-2)− | 1.3512e-1(2.26e-2)− | 1.8196e-1(4.41e-2)− | 2.1519e-1(4.04e-2)− | 9.9234e-2(1.17e-2) |
|  | (20, 10) | 1.3420e-1(3.56e-2)− | 1.0760e-1(2.07e-2)− | 7.6663e-2(8.47e-3)− | 1.0469e-1(2.55e-2)− | 1.1902e-1(1.34e-2)− | 6.9168e-2(1.34e-2) |
| DF10 | (5, 10) | 8.5074e-2(1.35e-2)+ | 8.9613e-2(1.02e-2)+ | 2.5102e-1(4.11e-2)− | 7.4886e-2(1.12e-2)+ | 8.0877e-2(1.79e-2)+ | 1.1331e-1(4.35e-2) |
|  | (10, 10) | 7.6177e-2(7.38e-3)+ | 7.5711e-2(7.57e-3)+ | 1.2727e-1(3.34e-2)− | 7.0567e-2(1.21e-2)+ | 6.8774e-2(1.28e-2)+ | 7.6405e-2(2.58e-2) |
|  | (20, 10) | 7.1794e-2(9.82e-3)+ | 7.3375e-2(1.1f-2)≈ | 9.1274e-2(3.05e-2)− | 6.9955e-2(5.75e-3)+ | 7.0650e-2(1.28e-2)+ | 8.0568e-2(1.43e-2) |
| DF11 | (5, 10) | 5.7234e-2(1.46e-3)− | 5.8331e-2(2.06e-3)− | 6.9308e-2(9.75e-3)− | 6.0626e-2(1.21e-3)− | 7.4976e-2(3.19e-3)− | 5.6387e-2(1.02e-3) |
|  | (10, 10) | 5.5363e-2(2.27e-3)≈ | 5.5572e-2(1.28e-3)≈ | 5.7144e-2(3.16e-3)− | 5.6510e-2(9.84e-4)− | 6.9652e-2(1.86e-3)− | 5.5476e-2(2.01e-3) |
|  | (20, 10) | 5.4857e-2(1.15e-3)≈ | 5.4376e-2(1.02e-3)≈ | 5.4794e-2(1.32e-3)≈ | 5.5421e-2(1.17e-3)≈ | 6.6101e-2(2.37e-3)− | 5.4889e-2(1.26e-3) |
| DF12 | (5, 10) | 1.1471e-1(9.86e-3)≈ | 1.2524e-1(1.62e-2)− | 2.2940e-1(5.19e-2)− | 1.0725e-1(1.41e-2)+ | 1.4434e-1(2.68e-2)− | 1.0907e-1(1.75e-2) |
|  | (10, 10) | 9.1250e-2(1.09e-2)+ | 9.8196e-2(1.19e-2)≈ | 1.4998e-1(1.47e-2)− | 9.1923e-2(2.8.48e-3)+ | 1.2200e-1(2.84e-2)− | 1.0576e-1(1.95e-2) |
|  | (20, 10) | 9.0998e-2(1.08e-2)≈ | 8.9872e-2(1.20e-2)≈ | 9.6555e-2(1.02e-2)− | 8.8378e-2(1.33e-2)≈ | 9.3135e-2(9.55e-3)≈ | 8.8569e-2(1.32e-2) |
| DF13 | (5, 10) | 1.6295e-1(8.22e-3)− | 1.8830e-1(1.62e-2)− | 3.4057e-1(9.00e-2)− | 1.9125e-1(1.11e-2)− | 3.2097e-1(5.16e-2)− | 1.0708e-1(4.45e-3) |
|  | (10, 10) | 1.0456e-1(2.73e-3)− | 1.1006e-1(3.37e-3)− | 1.5492e-1(5.78e-2)− | 1.0856e-1(4.03e-3)− | 1.9289e-1(8.92e-2)− | 9.6815e-2(2.78e-3) |
|  | (20, 10) | 9.6409e-2(1.55e-3)− | 9.6610e-2(2.24e-3)− | 9.5809e-2(2.00e-3)− | 9.5702e-2(2.33e-3)− | 1.1391e-1(2.19e-2)− | 9.4029e-2(2.00e-3) |
| DF14 | (5, 10) | 4.2568e-1(3.82e-2)− | 4.1184e-1(3.74e-2)− | 2.8549e-1(2.53e-2)− | 3.8033e-1(5.02e-2)− | 4.5765e-1(5.22e-2)− | 1.8212e-1(2.00e-2) |
|  | (10, 10) | 4.9630e-1(4.52e-2)− | 4.6543e-1(3.46e-2)− | 2.1869e-1(3.12e-2)≈ | 4.4418e-1(3.61e-2)− | 5.2718e-1(4.34e-2)− | 1.9956e-1(3.24e-2) |
|  | (20, 10) | 9.3632e-1(6.50e-2)− | 9.5419e-1(7.85e-2)− | 4.8228e-1(4.11e-2)− | 9.6381e-1(7.44e-2)− | 1.1177e+0(5.84e-2)− | 4.4276e-1(6.62e-2) |
| +/−/≈ |  | 5/25/12 | 3/27/12 | 1/28/13 | 4/26/12 | 2/35/5 |  |

## 5.2 Ablation Study

In this section, we design an ablation experiment to provide a deeper insight of the performance obtained by the proposed KAEP strategy and demonstrate the efficacy of the combination of centroid-based and KAE-based prediction. Especially, We designed two different types of autoencoding evolutionary search methods to validate that the kernelized autoencoding model is better than the linear autoencoding model used in dynamic multi-objective optimization.

Specifically, the design of the comparative algorithms in the ablation study is as follows.

1) CP-NSGA-II: we use NSGA-II as the baseline optimization method and then only use the centroid to predict the entire population, where the centroid is calculated by Formula (10) same as KAEP-NSGA-II.

2) KAE-NSGA-II: we only use KAE to generate the entire initial population without prediction based on the centroid when the dynamic occurs.

3) AEa-NSGA-II: we use NSGA-II as the baseline optimization method and then only use the linear autoencoding to generate the entire population.

| Problem | $(\tau_t, n_t)$ | CP-NSGA-II | AEa-NSGA-II | KAE-NSGA-II | AEb-NSGA-II | KAEP-NSGA-II |
|---|---|---|---|---|---|---|
| DF1 | (5, 10) | 5.8917e-2(5.44e-3)− | 3.6114e+0(4.63e-2)− | 8.5896e-1(1.03e+0)− | 5.3513e-2(9.55e-3)≈ | 4.8504e-2(7.27e-3) |
|  | (10, 10) | 1.6336e-2(9.33e-4)− | 2.7873e+0(5.38e-2)− | 9.6345e-1(8.58e-1)− | 1.4451e-2(9.51e-4)− | 1.4020e-2(1.08e-3) |
|  | (20, 10) | 6.5478e-3(1.45e-4)− | 1.6197e+0(5.18e-2)− | 6.0503e-1(4.02e-1)− | 6.5734e-3(1.77e-4)− | 6.4373e-3(1.88e-1) |
| DF2 | (5, 10) | 2.6448e-1(1.82e-2)− | 3.3883e+0(6.06e-2)− | 2.1284e+0(1.01e+0)− | 1.7461e-1(1.76e-2)+ | 2.0479e-1(2.22e-2) |
|  | (10, 10) | 1.5845e-1(1.18e-2)− | 2.4764e+0(5.65e-2)− | 1.3758e+0(6.86e-1)− | 9.3364e-2(1.15e-2)+ | 1.1429e-1(1.24e-2) |
|  | (20, 10) | 6.8013e-2(8.52e-3)− | 1.3054e+0(7.32e-2)− | 3.3565e-1(1.77e-1)− | 4.1448e-2(6.52e-3)+ | 5.4703e-2(8.63e-3) |
| DF3 | (5, 10) | 2.5972e-1(4.62e-2)− | 2.9436e+0(6.66e-2)− | 7.5105e-1(3.13e-1)− | 2.4408e-1(5.33e-2)− | 1.8831e-1(6.27e-2) |
|  | (10, 10) | 1.9903e-1(3.15e-2)− | 1.8369e+0(6.84e-2)− | 6.4261e-1(3.11e-1)− | 2.0305e-1(4.07e-2)− | 1.8016e-1(3.61e-2) |
|  | (20, 10) | 1.7251e-1(3.94e-2)− | 8.1584e-1(6.76e-2)− | 3.8158e-1(9.40e-2)− | 1.6687e-1(3.97e-2)− | 1.2630e-1(4.90e-2) |
| DF4 | (5, 10) | 1.2337e-1(6.51e-3)− | 1.6808e+0(1.34e-1)− | 1.6994e-1(2.49e-2)− | 1.1605e-1(9.57e-3)− | 1.0577e-1(4.53e-3) |
|  | (10, 10) | 8.8324e-2(4.14e-3)+ | 8.0306e-1(1.00e-1)− | 1.2139e-1(1.17e-2)− | 8.9822e-2(4.36e-3)+ | 9.2889e-2(3.20e-3) |
|  | (20, 10) | 8.8887e-2(2.60e-3)+ | 2.3888e-1(1.88e-2)− | 9.2531e-2(5.15e-3)≈ | 9.0088e-2(1.88e-3)+ | 9.1994e-2(1.89e-3) |
| DF5 | (5, 10) | 4.3362e-2(9.74e-3)+ | 2.9490e+0(8.12e-2)− | 2.8774e-1(4.20e-1)− | 1.1442e-1(6.12e-2)− | 5.0479e-2(7.87e-3) |
|  | (10, 10) | 1.2411e-2(2.55e-4)+ | 1.8343e+0(6.78e-2)− | 2.1052e-1(1.41e-1)− | 2.2434e-2(5.34e-2)− | 1.4908e-2(1.17e-3) |
|  | (20, 10) | 6.6600e-3(1.47e-4)+ | 7.3101e-1(7.91e-2)− | 7.9319e-2(7.12e-2)− | 7.1385e-2(5.98e-3)− | 7.7099e-3(3.87e-4) |
| DF6 | (5, 10) | 4.3137e+0(9.28e-1)≈ | 2.3750e+1(2.92e+0)− | 1.0988e+1(7.50e+0)− | 4.0218e+0(6.95e-1)+ | 5.9864e+0(7.60e-1) |
|  | (10, 10) | 2.8690e+0(1.16e+0)≈ | 8.8704e+0(1.39e+0)− | 4.0589e+0(9.23e-1)− | 2.0278e+0(6.13e-1)+ | 2.6626e+0(7.16e-1) |
|  | (20, 10) | 2.8180e+0(1.48e+0)≈ | 2.4237e+0(4.63e-1)− | 1.7530e+0(7.86e-1)≈ | 1.4461e+0(4.76e-1)≈ | 1.4194e+0(5.42e-1) |
| DF7 | (5, 10) | 3.3303e-1(3.95e-2)− | 3.4285e+0(2.86e-1)− | 4.3281e-1(5.63e-2)− | 3.2022e-1(3.97e-2)− | 2.1638e-1(7.16e-2) |
|  | (10, 10) | 3.2869e-1(2.79e-2)− | 2.1632e+0(3.84e-1)− | 4.1253e-1(3.19e-2)− | 3.0916e-1(3.55e-2)− | 2.0550e-1(6.94e-2) |
|  | (20, 10) | 3.0363e-1(5.61e-2)− | 6.6143e-1(1.22e-1)− | 3.9020e-1(3.43e-2)− | 3.0502e-1(4.18e-2)− | 1.5929e-1(6.77e-2) |
| DF8 | (5, 10) | 1.5530e-2(1.76e-3)− | 2.6999e-1(2.98e-2)− | 4.1107e-1(2.23e-2)− | 1.4941e-2(2.00e-3)− | 1.4191e-2(1.57e-3) |
|  | (10, 10) | 1.1331e-2(8.67e-4)≈ | 1.8546e-1(2.51e-2)− | 4.4896e-1(4.61e-2)− | 1.1955e-2(6.72e-4)− | 1.1605e-2(9.08e-4) |
|  | (20, 10) | 1.0068e-2(5.98e-4)≈ | 1.0564e-1(1.83e-2)− | 5.6790e-2(1.20e-2)− | 9.9387e-3(5.95e-4)≈ | 9.6912e-3(5.14e-4) |
| DF9 | (5, 10) | 1.8817e+0(3.76e-1)− | 2.3048e+0(2.02e-1)− | 1.7423e+0(3.39e-1)− | 4.1253e-1(1.67e-1)− | 1.5494e-1(1.62e-2) |
|  | (10, 10) | 7.5277e-1(1.79e-1)− | 1.4079e+0(1.24e-1)− | 1.0377e+0(1.78e-1)− | 2.2639e-1(9.33e-2)− | 1.0314e-1(1.15e-2) |
|  | (20, 10) | 3.0284e-1(7.92e-2)− | 7.9765e-1(5.92e-2)− | 5.7071e-1(1.12e-1)− | 1.4284e-1(4.81e-2)− | 8.3176e-2(1.64e-2) |
| DF10 | (5, 10) | 1.7433e-1(4.93e-2)≈ | 2.3084e-1(5.50e-2)− | 1.5238e-1(1.72e-2)− | 2.0695e-1(4.57e-2)− | 1.5617e-1(3.20e-2) |
|  | (10, 10) | 1.4900e-1(3.93e-2)≈ | 1.9221e-1(5.51e-2)− | 1.3863e-1(4.22e-2)− | 1.5849e-1(2.86e-2)− | 1.2900e-1(2.44e-2) |
|  | (20, 10) | 1.2872e-1(3.51e-2)≈ | 1.6791e-1(3.74e-2)− | 1.3473e-1(4.21e-2)− | 1.4426e-1(5.38e-2)− | 1.0266e-1(1.02e-2) |
| DF11 | (5, 10) | 6.4984e-1(1.99e-3)− | 8.2741e+1(1.26e-2)− | 6.5145e-1(3.88e-3)− | 6.4586e-1(1.98e-2)− | 6.4138e-1(2.55e-3) |
|  | (10, 10) | 6.3899e-1(1.25e-3)− | 7.6626e+1(1.05e-2)− | 6.3754e-1(1.09e-3)− | 6.3713e-1(9.82e-4)− | 6.3353e-1(1.27e-3) |
|  | (20, 10) | 6.3301e-1(7.18e-4)− | 6.9740e+1(6.52e-3)− | 6.3140e-1(6.23e-4)− | 6.3203e-1(8.20e-4)− | 6.3068e-1(6.38e-4) |
| DF12 | (5, 10) | 1.9511e-1(2.01e-2)− | 7.6952e+1(3.17e-2)− | 3.8498e-1(8.66e-2)− | 1.8425e-1(3.45e-2)− | 1.3026e-1(6.83e-3) |
|  | (10, 10) | 1.3546e-1(6.52e-3)− | 7.2088e+1(1.78e-2)− | 2.5387e-1(5.49e-2)− | 1.7103e-1(1.08e-2)− | 1.0526e-1(4.59e-3) |
|  | (20, 10) | 1.0686e-1(3.63e-3)− | 6.4156e+1(2.63e-2)− | 1.3204e-1(2.76e-2)− | 9.8198e-2(2.22e-3)− | 9.1452e-2(2.51e-3) |
| DF13 | (5, 10) | 2.0156e-1(3.14e-2)− | 4.5048e+0(1.09e-1)− | 2.6976e-1(9.97e-2)− | 2.0289e-1(1.29e-2)− | 1.6702e-1(7.42e-3) |
|  | (10, 10) | 1.2718e-1(2.82e-3)− | 3.0269e+0(1.26e-1)− | 1.7923e-1(3.09e-2)− | 1.2890e-1(3.77e-3)− | 1.2406e-1(3.41e-3) |
|  | (20, 10) | 1.1089e-1(1.84e-3)≈ | 1.3649e+0(1.20e-1)− | 1.2709e-1(1.45e-2)− | 1.1004e-1(1.52e-3)≈ | 1.1008e-1(1.34e-3) |
| DF14 | (5, 10) | 2.0874e-1(7.29e-2)− | 2.4108e+0(6.40e-2)− | 3.0811e-1(3.75e-2)− | 4.8200e-1(1.11e-1)− | 9.1125e-1(2.18e-2) |
|  | (10, 10) | 1.5062e-1(4.68e-2)≈ | 1.5665e+0(6.40e-2)− | 2.2470e-1(3.29e-2)− | 3.0625e-1(9.45e-2)− | 1.5027e-1(1.93e-2) |
|  | (20, 10) | 1.2703e-1(3.63e-2)≈ | 6.6960e+1(6.53e-2)− | 1.4884e-1(1.36e-2)≈ | 2.3392e-1(6.86e-2)− | 1.3434e-1(2.02e-2) |
| +/−/≈ |  | 5/26/11 | 0/42/0 | 0/36/6 | 5/32/ |  |

4) AEb-NSGA-II: we combine the linear autoencoding model and the centroid to generate half of the population each, similar to KAEP-NSGA-II.

The statistical results of the mean and standard deviations of MIGD values are presented in Table V, KAEP-NSGA-II perfroms significantly better than CP-NSGA-II and AEb-NSGA-II in most cases forsolving DF test problems. Specifically, the proposed KAEP-NSGA-II achieves 26 out of 42 better MIGD values compared CP-NSGA-II, implying that the kernelized auto-encoding evolutionary search and its combination with centroid-based prediction strategy indeed assist the population to search for higher quality optimal solutions. For AEb-NSGAII, the proposed KAEP-NSGA-II achieves 32 out of 42 better MIGD values. To some extent, it is means that the kernlized autoencoding model can better help the algorithm achieve better performance than the linear autoencoding model, based on a same SMOEA (i.e., NSGA-II). Meanwhile, according to the results in Table V, we can see that using only the kernelized autoencoding and the linear autoencoding to generate the entire population did not yield good results, but KAE-NSGAII performs better than AEa-NSGA-II on most of the test problems in terms of MIGD values.

# 6 Conclusion

In this article, we propose a method to solve the DMOPs by using a kernelized autoencoding evolutionary search approach. The approach tracks the movement of the POS as the evolutionary search progresses online. In summary, our prediction strategy consists of two components: prediction via the kernelized autoencoding and a simple centroid of the elite solutions obtained previous dynamic environments. We derive a kernel autoencoder that has a closed-form solution for predicting the movement of the POS. The proposed prediction method is learned from the elite solutions found during the dynamic optimization process, which enables more accurate prediction of the POS's movement. Furthermore, we calculate the centroid of nondominated solutions obtained from previous two historical environments and use it to predict the moving of POS in new environment. Comprehensive empirical studies have been conducted on 14 complex multi-objective benchmarks, and the statistical results have demonstrated that the proposed KAEP strategy enhance the ability of the SMOEAs to handle dynamic multi-objective optimization problem. Additionally, an ablation study has shown that the kernelized autoencoding can significantly improve the performance of the prediction-based DMOEA for solving DMOPs.

In the future, we would like to conduct a comprehensive investigation into the impact of various kernel functions on the effectiveness of KAEP in solving optimization problems of diverse natures. Additionally, we will aim to explore the more efficient methods to leverage cutting-edge autoencoders for nonlinear mapping in more randomly environmental changes.

## Acknowledgment

This work was supported in part by the Natural Science Foundation of China (Grant No. 62276224), in part by the Natural Science Foundation of Hunan Province, China (Grant No. 2022JJ40452), and in part by the General Project of Hunan Education Department (Grant No. 21C0077).